\documentclass{article}

\PassOptionsToPackage{numbers, sort&compress}{ICLR2026/natbib}
\usepackage[section]{placeins}
\usepackage{enumitem}
\usepackage{hyperref}
\usepackage{url}
\usepackage{ICLR2026/iclr2026_conference}
\usepackage{times}
\iclrfinalcopy
\usepackage{amsmath,amsfonts,bm}

\def\eqref#1{equation~\ref{#1}}
\def\1{\bm{1}}

\DeclareMathAlphabet{\mathsfit}{\encodingdefault}{\sfdefault}{m}{sl}
\SetMathAlphabet{\mathsfit}{bold}{\encodingdefault}{\sfdefault}{bx}{n}

\usepackage[utf8]{inputenc} % allow utf-8 input
\usepackage[T1]{fontenc}    % use 8-bit T1 fonts
\usepackage{hyperref}       % hyperlinks
\usepackage{url}            % simple URL typesetting
\usepackage{booktabs}       % professional-quality tables
\usepackage{amsfonts}       % blackboard math symbols
\usepackage{nicefrac}       % compact symbols for 1/2, etc.
\usepackage{microtype}      % microtypography
\definecolor{citecolor}{HTML}{0071BC}
\definecolor{linkcolor}{HTML}{D32F2F}%{ED1C24}
\definecolor{cellcolor}{HTML}{E3F2FD}
\definecolor{red}{HTML}{D32F2F}
\definecolor{magenta}{HTML}{D81B60}

\usepackage{amsmath}
\usepackage{amssymb}
\usepackage{mathtools}
\usepackage{amsthm}

\usepackage[capitalize,noabbrev]{cleveref}

\theoremstyle{plain}

\theoremstyle{definition}

\theoremstyle{remark}

\usepackage[textsize=tiny]{todonotes}
\usepackage{microtype}
\usepackage{graphicx}
\usepackage{subfigure}
\usepackage{booktabs} % for professional tables

\usepackage{pgfplots}
\pgfplotsset{compat = newest}
\usepackage{multirow}

\usepackage{framed}
\usepackage{tcolorbox}
\usepackage{caption}
\usepackage{ICLR2026/algorithm}
\usepackage{mathrsfs}
\usepackage{mathtools}
\usepackage{subcaption}
\usepackage{tcolorbox}
\usepackage{listings}
\usepackage{makecell}
\usepackage{colortbl}
\usepackage{color}
\usepackage{wrapfig}
\usepackage{cancel}
\usepackage{soul,xcolor}
\usepackage{pifont}
\usepackage{tikz}
\usepackage{algpseudocode}
\usetikzlibrary{shapes, positioning, arrows.meta, calc, decorations.pathmorphing, quotes}
\usepackage{todonotes}
\usepackage{booktabs, multirow, siunitx, caption}
\usepackage{xcolor}
\definecolor{myred}{RGB}{220, 38, 38}   % 鲜明红色
\definecolor{myblue}{RGB}{37, 99, 235}
\definecolor{mygreen}{RGB}{34,197,94}
\definecolor{DarkGreen}{rgb}{0.43, 0.68, 0.28}
\definecolor{NavyBlue}{RGB}{0,0,128}
\newcommand{\yuting}[1]{\textcolor{DarkGreen}{[Yuting: #1]}}

\usepackage{amsmath}
\usepackage{wrapfig}
\renewcommand{\cite}{\citep}

\makeatletter
\let\old@makefnmark\@makefnmark

\renewcommand{\@makefnmark}{%
    \hbox{\textsuperscript{\normalfont\textcolor{black}{\@thefnmark}}}%
}

\let\old@maketitle\@maketitle
\def\@maketitle{%
    \renewcommand{\@makefnmark}{%
        \hbox{\textsuperscript{\normalfont\textcolor{black}{\@thefnmark}}}%
    }%
    \old@maketitle
    \let\@makefnmark\old@makefnmark  % 恢复原始设置
}
\makeatother

\title{IDER: IDempotent Experience Replay for Reliable Continual Learning}
\usepackage{hyperref}
\usepackage{url}
\hypersetup{colorlinks=true, linkcolor=linkcolor, citecolor=citecolor,urlcolor=magenta}
\newcommand*\samethanks[1][\value{footnote}]{\footnotemark[#1]}

\author{%
    \quad\quad\quad\quad\quad \textbf{Zhanwang Liu}$^{1,}$\thanks{Equal contribution. This work was conducted at MIFA Lab (members from SJTU \& SII).}
    \quad
    \textbf{Yuting Li}$^{1,}$\samethanks[1]
    \quad
    \textbf{Haoyuan Gao}$^{1}$
    \quad
    \textbf{Yexin Li}$^{4}$\\[0.2cm]
    \quad\quad\quad\quad\quad\quad\quad \textbf{Linghe Kong}$^{1}$
    \quad
    \textbf{Lichao Sun}$^{3}$
    \quad
    \textbf{Weiran Huang}$^{1,2,}$\thanks{Correspondence to Weiran Huang (weiran.huang@outlook.com)}
    \\[0.3cm]
    $^1$ School of Computer Science, Shanghai Jiao Tong University \quad
    $^2$ Shanghai Innovation Institute\\[0.1cm]
    \quad \quad $^3$ Lehigh University \quad
    $^4$ State Key Laboratory of General Artificial Intelligence, BIGAI
}

\begin{document}

\maketitle

\begin{abstract}
%Catastrophic forgetting has been a major challenge in continual learning, where the model needs to learn new
%1.灾难性遗忘是什么2. CL方法可以解决3.CL can benifit from reliable prediction 4. 现有方法不能应用到setting上参数多与主流方法不兼容5.转折，引入id来实现reliable prediction，介绍id（从句）6. 方法的步骤7.实验结果 We conduct, observe 8. suggest Id 作用
    Catastrophic forgetting, the tendency of neural networks to forget previously learned knowledge when learning new tasks, has been a major challenge in continual learning (CL). To tackle this challenge, CL methods have been proposed and shown to reduce forgetting. Furthermore, CL models deployed in mission-critical settings can benefit from uncertainty awareness by calibrating their predictions to reliably assess their confidences. However, existing uncertainty-aware continual learning methods suffer from high computational overhead and incompatibility with mainstream replay methods. To address this, we propose idempotent experience replay (IDER), a novel approach based on the idempotent property where repeated function applications yield the same output. Specifically, we first adapt the training loss to make model idempotent on current data streams. In addition, we introduce an idempotence distillation loss. We feed the output of the current model back into the old checkpoint and then minimize the distance between this reprocessed output and the original output of the current model. This yields a simple and effective new baseline for building reliable continual learners, which can be seamlessly integrated with other CL approaches. Extensive experiments on different CL benchmarks demonstrate that IDER consistently improves prediction reliability while simultaneously boosting accuracy and reducing forgetting. Our results suggest the potential of idempotence as a promising principle for deploying efficient and trustworthy continual learning systems in real-world applications.
     Our code is available at \url{https://github.com/YutingLi0606/Idempotent-Continual-Learning}.

    \end{abstract}

\section{Introduction}

Deep learning has achieved impressive success across various domains. However, a static batch setting where the training data of all classes can be accessed at the same time is essential for attaining good performance~\cite{le2015tiny,rebuffi2017icarl}. In many real-world deployments, data arrive sequentially and previously seen samples cannot be fully retained due to storage or privacy constraints. This makes it a major challenge because neural networks tend to rapidly forget previously learned knowledge when trained on new tasks, which is a phenomenon known as catastrophic forgetting~\cite{mccloskey1989catastrophic}. 

To address this challenge, continual learning (CL) is proposed to enable models to accumulate knowledge as data streams arrive sequentially. Among valid CL strategies, rehearsal-based approaches are popular as they are simple and efficient. They~\cite{chaudhry2019continual,wu2019large,buzzega2020dark,caccia2021new,boschini2022class} address this by storing a small, fixed-capacity buffer of exemplars from previous tasks and replaying them when training on new task, thereby regularizing parameter updates and mitigating catastrophic forgetting. %Recent rehearsal-based approaches ~\cite{gu2023preserving,sarfrazsemantic} focus primarily on improving average accuracy while overlooking persistent overconfidence predictions of CL methods, which is more pronounced in CL where models tend to be biased towards recent tasks~\cite{arani2022learning,cai2024our}. 
Despite strong average accuracy, CL methods are often poorly calibrated and over-confident, a problem exacerbated by recency bias toward new tasks~\cite{arani2022learning}.
Thus, this undermines the broader deployment of CL models in real-world settings~\cite{li2024sure}, especially in safety‑critical domains (healthcare, transport, etc.)~\cite{lecun2022path}. CL models deployed in these domains can benefit from uncertainty awareness by calibrating their predictions to reliably assess their confidences~\citep{jha2024clap4clip}. To tackle this issue, ~\citet{jha2023npcl} propose neural processes based CL method (NPCL). However, it causes non-negligible parameter growth and exhibits incompatibility with logits-based replay methods due to the stochasticity in the posterior induced by Monte Carlo sampling. Motivated by these limitations, we aim for a lightweight and compatible principle for reliable CL methods. 
\begin{comment}
\begin{table}[htbp]
\centering
\caption{Model calibration errors on different datasets}
\label{tab:model_performance}
\begin{tabular}{lccc}
\toprule
\multirow{2}{*}{\textbf{Model}} & \textbf{CIFAR-10} & \textbf{CIFAR-100} & \textbf{CIFAR-100} \\
 & Buffer 500 & Buffer 500 & Buffer 2000 \\
\midrule
ER & 32.69 & 64.48 & 45.64 \\
DER++ & 17.02 & 17.32 & 16.12 \\
ER+ID & 12.36 & 12.87 & 13.65 \\
BFP & 9.83 & 9.28 & 11.93 \\
BFP+ID & \textbf{8.63} & \textbf{8.93} & \textbf{8.83} \\
\bottomrule
\end{tabular}
\end{table}
\end{comment}

We draw inspiration from idempotence, a mathematical property that arises in algebra. An operator is idempotent if applying it multiple times yields the same result as applying it once, formally expressed as $f(f(x)) = f(x)$. It can be used in deep learning by recursively feeding the model’s predictions back as inputs, allowing the model to refine its outputs~\cite{shocher2023idempotent,durasov2024zigzag}. ~\citet{durasov2024enabling} empirically demonstrate that if a deep network $f$ takes as input a vector $x$ and a second auxiliary variable that can either be the ground truth label $y$ corresponding to $x$ or a neutral uninformative signal $0$ and is trained so that $f(x, 0) = f(x, y) = y$, then the distance $||f(x, f(x, 0)) - f(x, 0)||$ correlates strongly with the prediction error. What if we actively minimize this distance of buffer data when we learn new tasks in CL settings? Could we project outputs into the stable manifold where instances are mapped to themselves to prevent predictive distribution drift? 
%We hypothesize that enforcing idempotence for continual learning stabilizes the predictive distribution and reduces decision-boundary drift, thereby improving both accuracy and calibration.

\begin{figure}[t]
    \centering
    \includegraphics[width=\linewidth]{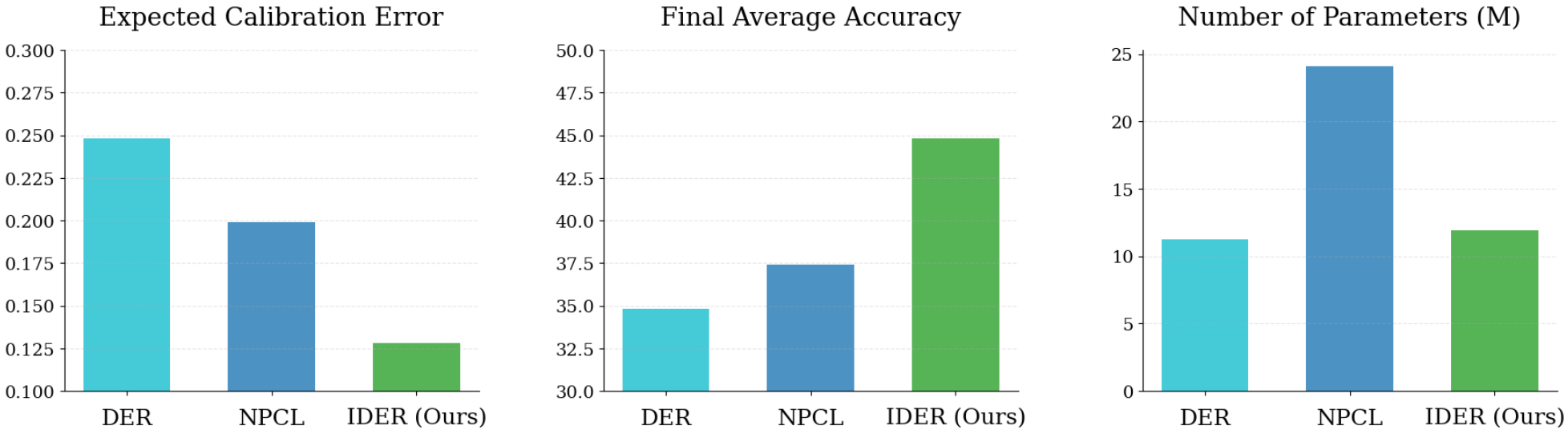}
    \caption{ We propose the IDER method, which can be directly applied to many recent rehearsal-based continual learning methods, resulting in less calibration error and significant improvements in FAA with less parameter growth compared with NPCL.}
    \label{fig:T}
\end{figure}

Thus, we propose an Idempotent Experience Replay (IDER) inspired by Idempotence, a simple and effective method that enforces idempotence for CL models when learning new tasks. We demonstrate that enforcing idempotence enables model to make more reliable predictions while reducing catastrophic forgetting. Both combined with naive rehearsal-based method experience replay (ER)~\cite{riemer2019learning}, compared with NPCL, our approach achieves lower calibration error evaluated by Expected Calibration
Error (ECE)~\cite{guo2017calibration}, higher accuracy, and requires smaller parameter numbers, as is shown in Figure~\ref{fig:T}.

More specifically, IDER integrates two components to enforce idempotence for CL models. Firstly, we adapt the training loss to train the current model to be idempotent with data from the current task.  Secondly, we introduce idempotence distillation loss for both buffer data and the current data stream to enforce idempotence between last task model checkpoint $f_{t-1}$ and current model $f_{t}$. We verify that incorporating the current data stream into idempotence achieves further performance improvements, suggesting that idempotence can help preserve model distribution, thereby mitigating decision boundary drift over time.

This yields a simple method that only requires two forward passes of the model almost without additional parameters. Our approach can be integrated into existing CL methods and experiments show that this simple change boosts both prediction reliability and final accuracy by a large margin. Especially on the Split-CIFAR10 dataset, enforcing idempotence improves the baseline method ER~\cite{riemer2019learning} by up to 26$\%$, achieving state-of-the-art class incremental learning accuracy. Through extensive empirical validation on challenging generalized class-incremental learning ~\cite{mi2020generalized,sarfrazsemantic}, we demonstrate that this simple and powerful principle improves the reliability of predictions while mitigating catastrophic forgetting in real-world scenarios.

The contributions of this paper can be summarized as follows:

\begin{itemize}[itemsep=0pt,topsep=0pt,leftmargin=0.5cm]
%   \item We introduce a new perspective for continual learning, that continual learning methods should not focus only on mitigating catastrophic forgetting but also ensure reliable predictions and avoid overconfidence..  
    \item We propose a novel framework for continual learning based on the idempotent property, which is a simple and robust method. Our method demonstrates that fundamental mathematical properties can be effectively utilized to address catastrophic forgetting for CL. 
    
    \item We show that IDER can be easily integrated into other state-of-the-art methods, leading to more reliable predictions with comparable performance.

    \item Extensive experiments on several benchmarks demonstrate that our approach achieves strong performance in both mitigating catastrophic forgetting and making reliable predictions.
 
\end{itemize}

%Our work establishes a new paradigm for addressing catastrophic forgetting in continual learning, showing how fundamental mathematical principles can be leveraged to create more effective and reliable learning systems. The simplicity and theoretical grounding of our approach make it particularly attractive for practical applications while opening new directions for future research in continual learning.

\section{Related Work}

\paragraph{Continual Learning.}
The goal of continual learning (CL) is to achieve the balance between learning plasticity and memory stability~\cite{wang2024comprehensive}. 
%One major challenge is called catastrophic %forgetting~\cite{mccloskey1989catastrophic},where %adaptation to changes in the distribution of input %data will cause significantly loss of previously %learned knowledge. 
Approaches in CL can be divided into three main categories. Regularization-based methods primarily rely on regularization loss to penalize changes in parameter space of the model ~\cite{farajtabar2020orthogonal,kirkpatrick2017overcoming}. Rehearsal-based Method ~\cite{chaudhry2019continual} use a memory buffer to store task data and replay them during new task training. Architecture-based methods ~\cite{rusu2016progressive,wang2022learning} incrementally expand the network to allocate distinct parameters for preserving each task's knowledge. Recently, CL has also been studied in more practical settings such as continual post-training of multi-modal LLMs~\cite{weifirst,xu2025llava,yan2026livemedbench}, continual offline RL~\cite{hu2025tackling,hu2025continual}, and federated continual learning ~\cite{yao2024variational}. Among them, Rehearsal-based methods are general in various CL scenarios and can be naturally combined with knowledge Distillation (KD) techniques. 
%Parameter regularization approaches like EWC~\cite{kirkpatrick2017overcoming} and SI~\cite{zenke2017continual} introduce new loss as regularization terms by constraining weight updates largely to prevent forgetting. 

The baseline Experience Replay (ER) ~\cite{riemer2019learning} mixes the current task data with stored samples from past tasks in the memory buffer during training.
%iCaRL~\cite{rebuffi2017icarl} is the first replay based method to perform KD, by minimizing the loss function that consists of classification and distillation terms. %LUCIR~\cite{hou2019learning} encourages the orientation of features extracted by current network to be similar to those by the old model. 
%It also uses cosine normalization to reduce the bias for new classes. 
%BiC~\cite{wu2019large} assumes the last fully connected layer is biased which casuses catastrophic forgetting.
%To correct the bias caused by the last fully connected layer, BiC~\cite{wu2019large} uses the linear model with the balanced validation set from old and new classses. 
DER ~\cite{buzzega2020dark} store old training samples together with their logits and preserve the old knowledge by matching the saved logits with logits obtained by current model. Its improved version XDER~\cite{boschini2022class} improves performance at the sacrifice of computational costs due to sophisticated mechanisms. CLSER~\cite{arani2022learning} introduce a fast module for plastic knowledge and a slow learning module for stable knowledge. BFP~\cite{gu2023preserving} uses a
learnable linear layer to perform knowledge distillation in the feature space. SCoMMER~\cite{sarfraz2023sparse} and SARL~\cite{sarfrazsemantic} enforces sparse coding for efficient representation
learning. Neural Processes for Continual Learning (NPCL)~\cite{jha2023npcl} explore uncertainty-aware CL models using neural processes (NPs). 
%Unlike previous studies, we explore the idempotence in continual learning, which has never been studied before.

\paragraph{Idempotence in Deep Learning.}

Idempotence is a property of a function whereby the result of applying the function once is the same as applying it multiple times in sequence.
Recent work has explored the application of idempotence in deep learning. In particular, it is defined that the results obtained by the model will not change when applying the model multiple times in practice ($f(f(x)) = f(x)$). 
The Idempotent Generative Network (IGN)~\cite{shocher2023idempotent} 
firstly proposes this idea in deep learning for generative modeling and it has the capability of producing robust outputs in a single step.
Another work  ZigZag ~\cite{durasov2024zigzag} introduces idempotence in neural networks for the measuring uncertainty, which is based on IterNet~\cite{durasov2024enabling}. IterNet  proves that for iterative architectures, which use their own output as input, the convergence rate of their successive outputs is highly correlated with the accuracy of the value to which they converge. %ZigZag introduces a sample free uncertainty estimation method with idempotence property. 
ZigZag recursively feeds predictions back as inputs, measuring the distance between successive results. A small distance indicates high confidence, while a large one signals uncertainty or out-of-distribution (OOD) data.
%This approach offers computational advantages over sampling-based uncertainty methods~\cite{gal2016dropout,lakshminarayanan2017simple} by avoiding multiple forward passes.
Recent work ITTT~\cite{durasov20243} combines idempotence with Test-Time Training. These works proves the potential of idempotence in deep learning while these works are based on static batch learning.
%With idempotent loss, it projects the internal representation of the input onto the training distribution, which improves robustness on corrupted and OOD data  by refining on a single instance each time at test-time.

\section{Method}
\label{sec:method}

\begin{figure}[t]
    \centering
    \includegraphics[width=0.8\linewidth]{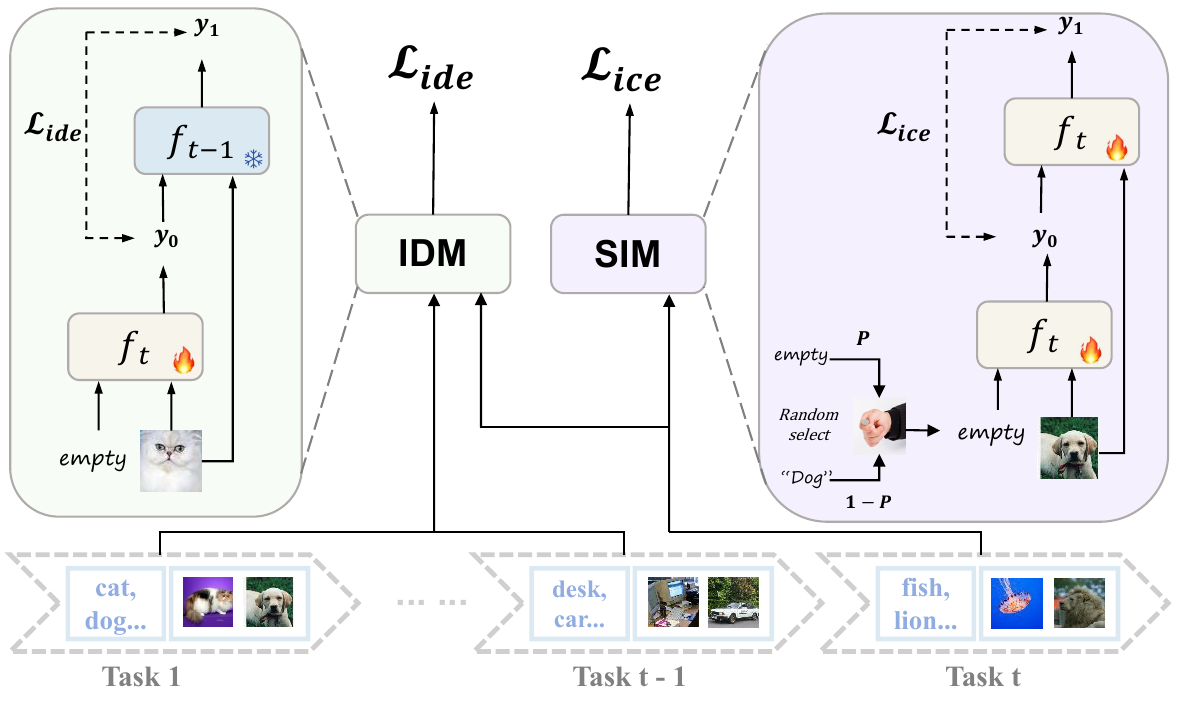}
    \caption{Overall framework of Idempotent Experience Replay (IDER). Our method consists of two modules for continual learning: (1) Standard Idempotent Module that trains current model idempotent with data from the current task. (2) Idempotent Distillation Module that  enforce the current model to become idempotent with respect to the last task model checkpoint, utilizing data from both the current task and buffer memory. IDER can be integrated into existing CL approaches to make reliable predictions while mitigate catastrophic forgetting.}
    \label{fig:method}
\end{figure}

In this section, we deliver details of the proposed IDER. We first define both class-incremental learning and generalized class-incremental learning settings. Then, we elaborate on how to introduce idempotence in continual learning. Finally, we introduce the overall objective. An overview of IDER is depicted in Figure~\ref{fig:method}.

\subsection{Problem Definition}

Continual learning (CL) aims to develop models that learn from a stream of tasks while preserving previously acquired knowledge. In this paper, we focus on typical class-incremental learning (CIL) and generalized class-incremental learning. Generalized class-incremental learning (GCIL) is more close to real-world incremental learning. The key GCIL properties can be summarized as follows: (i) the
number of classes across different tasks is not fixed; (ii) classes shown in prior tasks could reappear
in later tasks; (iii) training samples are imbalanced across different classes in each task.

In a typical class-incremental learning setting, a model $f$ is trained on sequential tasks $T={\mathcal{T}_1,\mathcal{T}_2,...,\mathcal{T}_{t}}$ Each task $\mathcal{T}$ consists of data points and these data points are unique within each task, which means $\mathcal{T}_{t} = \{(x_i, y_i)\}_{i=1}^{N_t}$ and $\mathcal{T}_{i} \cap \mathcal{T}_{j} = \emptyset$. The optimization objective is to minimize the overall loss over all the tasks: 
\begin{equation}
    f^* = \arg\min_f  \sum_{i=0}^{t} \mathbb{E}_{(x, y) \sim \mathcal{T}_{t}} \left[ \mathcal{L}(f(x), y) \right],
\end{equation}
where $L$ is the loss function for the tasks and $y$ is the ground truth for $x$. However, in the continual setting, only the data from current task $\mathcal{T}_{t}$  are available and the model should preserve the previous knowledge from the tasks before$\mathcal{T}_1,...,\mathcal{T}_{t-1}$. As a result, additional memory buffer or additional regularization term $\mathcal{L}_{R}$ may be chosen to avoid catastrophic forgetting and the actual objective on the current task should be:
\begin{equation}
    f^* = \arg\min_f  [ \mathbb{E}_{(x, y) \sim \mathcal{T}_{t}\cup \mathcal{M}} \left[ \mathcal{L}(f(x), y) \right]+\mathcal{L}_{R}].
    \label{eq:cl loss}
\end{equation}
where $\mathcal{M}$ stands for the memory buffer to store the data from previous tasks.

\subsection{Modified Architecture}

\begin{wrapfigure}{r}{0.4\textwidth}
    \vspace{-6pt}
    \centering
    %\captionsetup{font=small}
    \includegraphics[width=0.4\textwidth]{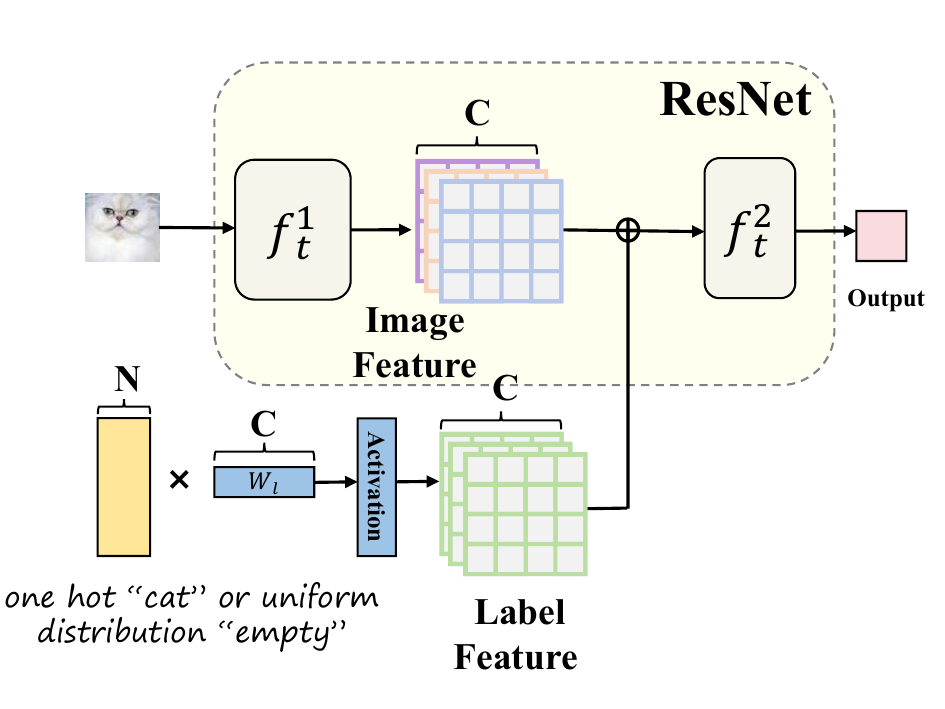}
    \caption{Modified Architecture. We modify the architecture of backbone(ResNet) and enable the model to accept two inputs.}
    \label{fig:arch}
    \vspace{-6pt}
\end{wrapfigure}
 To enable idempotence with respect to the second input, we modify the original backbone as shown in Figure~\ref{fig:arch}. We divide the backbone ResNet~\cite{he2016deep} as denoted $f_t$, into two parts $f_t^1$ and $f_t^2$ on the t-th task. 
 The second input (either a one-hot vector $y$ or a uniform distribution over all classes standing for ``empty'' input $\mathbf{0}$) is first transformed into a label feature vector. This is achieved by a linear layer with an output dimension that matches the dimensions of $f_t^1$'s output, followed by a LeakyReLU activation function. The image first is processed by $f_t^1$ to produce an intermediate feature map. The label feature is then added to this intermediate feature map, which is fed into $f_t^2$. The output of $f_t^2$, which is the logits for target classes, can work as the second input for model after softmax normalization. In this way, the backbone can accept two inputs and achieve idempotence with respect to the second argument after training.

\subsection{Standard Idempotent Module:Training the Network Idempotent }
First, we rely on the model we train being idempotent. To achieve this, Standard Idempotent Module is used for training the model on data from the current task. Following ~\citet{durasov2024enabling,durasov2024zigzag},  when learning new tasks, we minimize the loss which consists of two cross-entropy losses obtained by the logits from the first and second forward propagation of model and the ground truth $y$ :
\begin{equation}
    \mathcal{L}_{ice} = \sum_{(x, y) \in \mathcal{T}_{t}} [\mathcal{L}_{ce}(f_{t}(x,y^*),y) + \mathcal{L}_{ce}(f_{t}(x,f_{t}(x,y^*)), y)],
    \label{eq:ice loss}
\end{equation}  
where $\mathcal{T}_{t}$ is the current task and $y^*$ is the second input that is set to either the ground-truth one-hot vector $y$ or the neutral ``empty'' signal input $\mathbf{0}$. We randomly select $y$ with probability $1-P$ and the neutral ``empty'' signal input $\mathbf{0}$ with probability $P$. The empty signal $\mathbf{0}$ is defined as a uniform distribution over all classes.
%The concept of idempotence arises in algebra, which is first proposed and used in deep learning by  Idempotent Generative Network(IGN)~\cite{shocher2023idempotent}. IGN defines estimated manifold where data satisfy $f(x)=x$. Then it maps instances from a different distribution onto that estimated manifold, thus it can get that:
%\begin{equation}
%    \forall z \sim P_z, \quad f(f(z)) = f(z).
%\end{equation}
%Thus, the model  demonstrates potential in generalizing and reliability by projecting degraded
%or modified data back onto the estimated manifold. 

\begin{figure}[t]
    \centering
    \begin{minipage}[t]{0.48\textwidth}
        \centering        \includegraphics[height=6cm,width=\textwidth,keepaspectratio]{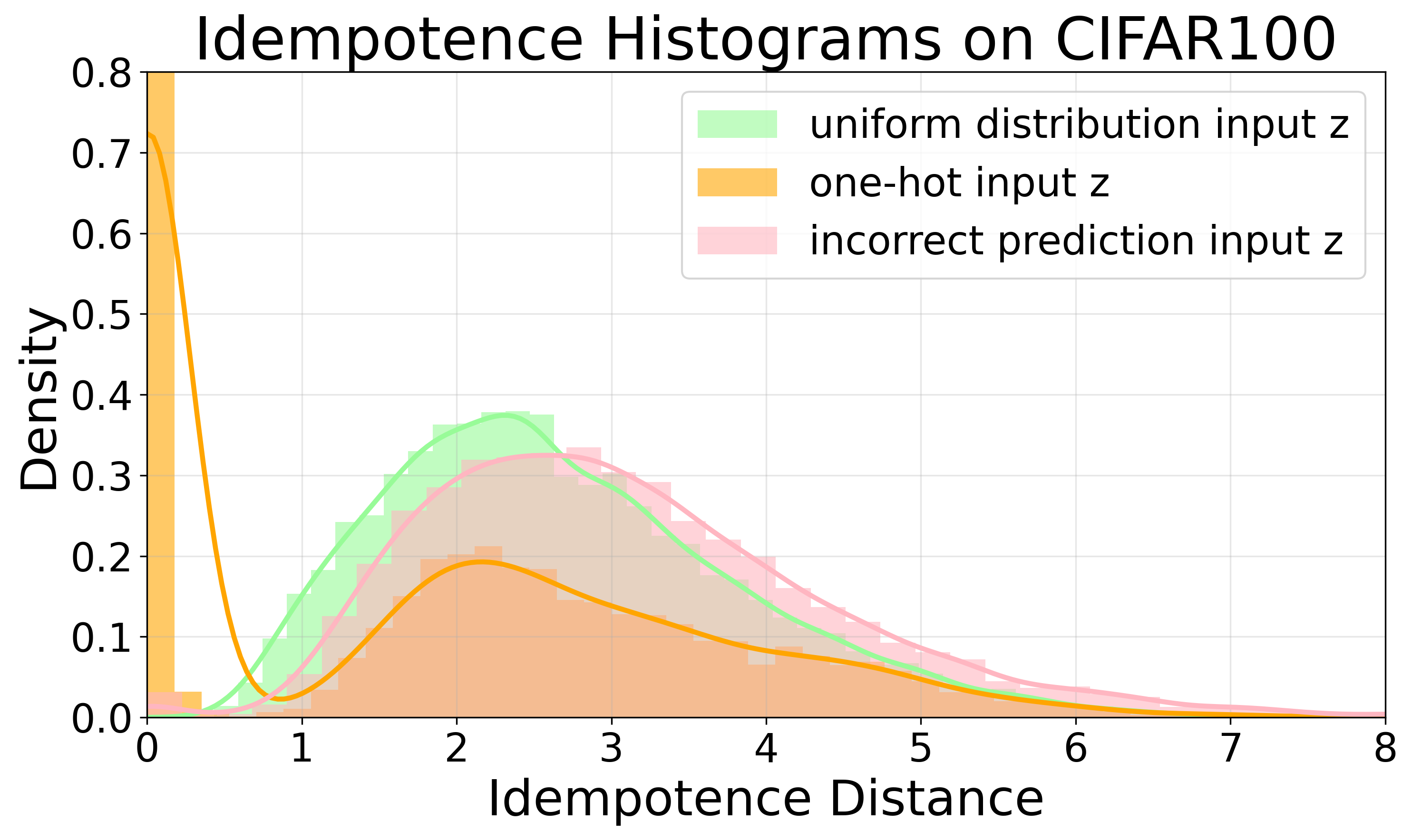}
        \caption{We plot the distribution of idempotence errors, measured by the distance $|f(x, f(x,z)) - f(x,z)|$. Inputs $x$ with second incorrect prediction input $z$ exhibit significantly larger idempotence errors. Thus, this distance can be used as a idempotent distillation loss.}
        \label{fig:Idempotence}
    \end{minipage}
    \hfill
     \begin{minipage}[t]{0.48\textwidth}
        \centering
        \includegraphics[height=6cm, width=\textwidth, keepaspectratio]{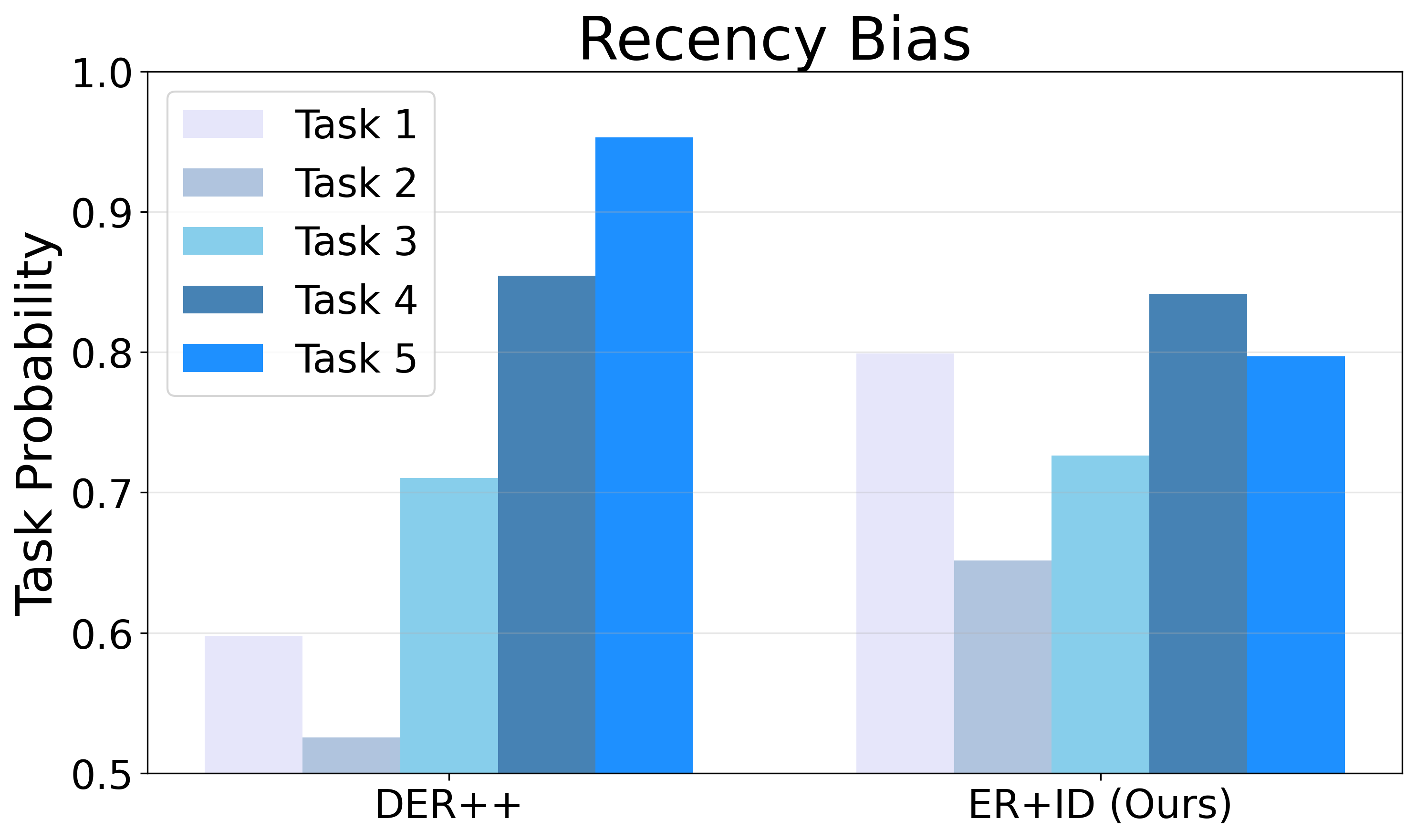}
        \caption{ Probability of predicting each task at the end of training for models trained on CIFAR-10 with 500 buffer size. Idempotent distillation loss effectively mitigates the bias to the recent tasks and provides a more uniform probability size across different tasks.}
        \label{fig:biasm}
    \end{minipage}
\end{figure}

By minimizing $\mathcal{L}_{ice}$, we can train the model idempotent with respect to the second argument, which can be obtained by:
\begin{equation}
 f_t(x, \mathbf{0}) \approx y, \quad 
 f_t(x, y)  \approx y,  \quad  f_t(x, f_t(x, \mathbf{0}) )\approx y
\implies f_t(x, f_t(x, \mathbf{0})) \approx f_t(x, \mathbf{0}).
\end{equation}
Thus, $f_{t}$ has been adjusted so that the model $f_t$ is as idempotent as possible for all $x$ in distribution. The model will map the data $(x,\mathbf{0})$ to the stable manifold $(x,y) : f(x,y=y)$. Fig.~\ref{fig:Idempotence} illustrates this in the case of a network trained on data from the first task on CIFAR-100. With different second input $y$, the idempotence distance distribution varies. The input which contains incorrect prediction input $y$ exhibits significantly larger idempotence errors. Thus, this distance can be used as a distillation loss for iterative prediction refinement to make reliable predictions. 
\subsection{Idempotent Distillation Module: Distilling The Network For Continual Learning }
In the CL setting, the model tends to have recency bias toward newly introduced classes, which negatively influences the performance and results in overconfidence predictions. Rehearsal-based methods suffer from this problem, as~\citet{wang2022learning} point out that when a new task is presented to the net, an asymmetry arises between the contributions of replay data and current examples to the weights updates: the gradients of new examples outweigh. To mitigate this issue, we require the model to maintain stable predictions on data from previous tasks even after parameter updates induced by new knowledge, as self-consistency
indicates that the network’s output is aligned with the learned in-distribution manifold and can
make reliable (well-calibrated) prediction. This condition can be translated into enforcing
idempotence. Thus, we propose to minimize idempotence distances to mitigate recency bias and prediction distribution drift in CL.
A naive way would be to minimize the loss function:
\begin{equation}
\mathcal{L}_{ide} = \sum_{(x, y) \in \mathcal{T}_{t}, M} \| f_{t}(x,\mathbf{0}) - f_{t}(x,f_{t}(x,\mathbf{0})) \|_2^2.
\end{equation}

However, this can produce undesirable side effects in CL settings. As $f_{t}$ has bias towards current data streams and $y_{0} = f_{t}(x, \mathbf{0})$ may be an incorrect prediction, minimizing $\|y0 - y1\|_2^2$ may cause $y_{1} = f_{t}(x, y_{0})$ to be pulled towards the incorrect $y_{0}$, thereby magnifying the error.

To address this, we keep the model checkpoint at the end of the last task $f_{t-1}$ together with the current trained model $f_{t}$. We then modify the idempotence distillation loss to be:
\begin{equation}
   \mathcal{L}_{ide} = \sum_{(x, y) \in \mathcal{T}_{t}, M} \| f_{t}(x,0) - f_{t-1}(x,f_{t}(x,0)) \|_2^2.
   \label{eq:ide loss}
\end{equation}
Thus, the first prediction $y_{0} = f_{t}(x, \mathbf{0})$ is computed as before, but the second one, $y_{1} = f_{t-1}(x, y_{0})$, is made using the last model checkpoint $f_{t-1}$ By updating only $f_{t}$ and keeping $f_{t-1}$ frozen, which preserves more previous knowledge and stable prediction distribution for buffer data, we ensure that $y_{0}$ is adjusted to minimize the discrepancy with $y_{1}$, without pulling $y_{1}$ towards an incorrect $y_{0}$.  Thus, we could prevent manifold expansion that would include mistaken outputs, focusing optimization solely on the first pass prediction and thereby avoiding error reinforcement.

This design achieves idempotence by ensuring that processing an input through the current model and then through the last model checkpoint yields a nearly identical output distribution. This self-consistency between the current model and the last model satisfy the requirement for idmpotence condition for sequential tasks in CL. What's more, $\mathcal{L}_{ide}$ can also serve as the distillation loss to mitigate catastrophic forgetting.   Unlike traditional distillation in ~\citet{buzzega2020dark}, which only aligns the final output probabilities, our method anchors the model’s representation to the stable manifold already learned by the frozen model, thereby maintaining balanced predictive performance across all tasks, as is shown in Figure~\ref{fig:biasm}. As a result, enforcing idempotence can help CL models mitigate catastrophic forgetting while make reliable preidctions.

\subsection{Overall Objective}

%The overall framework of our method is shown in Figure~\ref{fig:method}. It consists of two main components: Idempotent Distillation Module and Standard Idempotent Module. 

We introduce idempotence into an experience replay (ER) framework~\cite{riemer2019learning}, where we
keep a buffer $M$ storing training examples from old tasks. We keep the model checkpoint at the end of the last task $f_{t-1}$ together with the current trained model $f_{t}$. During continual learning,
the current model $f_{t}$ is trained on the batch from data stream of the current task $\mathcal{T}_{t}$ using the adapted training loss $\mathcal{L}_{ice}$ in Eq.~\ref{eq:ice loss}. We sample batch from $M$ and combine the current batch to compute idempotence distillation loss $\mathcal{L}_{ide}$ in Eq.~\ref{eq:ide loss}.

Meanwhile, we sample another batch from $M$ for experience replay. The experience replay loss $\mathcal{L}_{rep\text{-}ice}$ in ER is:

\begin{equation}
    \mathcal{L}_{rep\text{-}ice} = \sum_{(x, y) \in M} [\mathcal{L}_{ce}(f_{t}(x,y^*),y) + \mathcal{L}_{ce}(f_{t}(x,f_{t}(x,y^*)), y)].
\end{equation}  
%\paragraph{Idempotent Distillation Module} We train the model $f_{t}$ in the current task to satisfy the idempotent property $f_{t-1}(f_{t}(x)) = f_{t}(x)$ with the model checkpoint at the end of the last task $f_{t-1}$ with data from the fixed memory buffer. Therefore, the objective is:
%\begin{equation}
%   \mathcal{L}_{ide} = \sum_{(x, y) \in M} \| f_{t}(x,0) - f_{t-1}(x,f_{t}(x,0)) \|_2^2,
%\end{equation}
%where $0$ stands for the neutral "empty" signal input(uniform distribution over the one-hot label dimensions) and $M$ stands for memory buffer.

%During training on the task $t$, the model $f_t$ is optimized and the old model checkpoint $f_{t-1}$ remains fixed. 
%\paragraph{Standard Idempotent Module} Standard Idempotent Module is used for training the model with data from both the current task and the memory buffer. The loss consists of two cross-entropy losses obtained by the logits from the first and second forward propagation of model and the ground truth $y$ :
%\begin{equation}
%    \mathcal{L}_{ice} = \sum_{(x, y) \in \mathcal{T}_{t},M} [\mathcal{L}_{ce}(f_{t}(x,y^*),y) + \mathcal{L}_{ce}(f_{t}(x,f_{t}(x,y^*)), y)],
%\end{equation}  
%where $\mathcal{T}_{t}$ is current task and $y^*$ is the second input that is one-hot label $y$ with probability $P$ and the neutral "empty" signal input with probability $1-P$.

%The combination of two modules allows the model to achieve both effective knowledge transfer and stability while making stable predictions. The idempotent property acts as a natural regularizer, preventing the model from deviating too far from previously learned representations by satisfying idempotency for data from previous tasks while still allowing for the incorporation of new knowledge by making current model idempotent with current data stream. 
The total loss function used in IDER is the weighted sum of the losses above, formally:
\begin{equation}
    \mathcal{L}_{IDER} = \mathcal{L}_{ice} + \alpha \mathcal{L}_{ide} + \beta \mathcal{L}_{rep\text{-}ice}.
\end{equation} 
In addition, our method is simple and robust, which can be combined with other methods, such as BFP~\cite{gu2023preserving}, to achieve higher performances. Details are shown in the appendix.

\section{Experiments}
\noindent\textbf{Continual Learning Settings.}
We follow~\citet{gu2023preserving} and conduct experiments on state-of-the-art  rehearsal-based models in CIL setting which splits the dataset into a sequence of tasks, each containing a disjoint set of classes, while task identifiers are not available during testing. Following~\citet{sarfrazsemantic}, we also evaluate methods in the generalized class incremental learning (GCIL) setting~\cite{mi2020generalized} which is closest to the real-world scenario as the number of classes in each task is not
fixed, the classes can overlap and the sample size for each class can vary. 

\noindent\textbf{Evaluation Metrics.}
Following~\citet{buzzega2020dark,boschini2022class}, we use Final Average Accuracy (FAA) and Final Forgetting (FF) to reflect the performances of mitigating catastrophic.
We report well-established Expected Calibration Error (ECE)~\cite{guo2017calibration} to assess the reliability of continual learning methods. More details are shown in the appendix.
\begin{table}[t]
\centering
%\captionsetup{font=small}
\caption{Comparison of Final Average Accuracy (FAA) across different continual learning methods. All experiments are repeated 5 times with different seeds. Results for SARL~\cite{sarfrazsemantic} are from our implementation. The best results are highlighted in \textcolor{myblue}{blue}.The second best results are highlighted in \textcolor{mygreen}{green}.}
\renewcommand{\arraystretch}{1.4}
\resizebox{\textwidth}{!}{
\begin{tabular}{l cc cc cc}
\toprule
\multirow{2}{*}{\textbf{Method}} & \multicolumn{2}{c}{\textbf{CIFAR-10}} & \multicolumn{2}{c }{\textbf{CIFAR-100}} & \multicolumn{2}{c}{\textbf{Tiny-ImageNet}} \\
& Buffer 200 & Buffer 500 & Buffer 500 & Buffer 2000 & Buffer 500&Buffer 4000 \\
\midrule
Joint (upper bound) & \multicolumn{2}{c}{91.93\small{±0.29}}  & \multicolumn{2}{c}{71.15\small{±0.51}}  & \multicolumn{2}{c}{59.52\small{±0.33}} \\
\midrule
iCaRL~\cite{rebuffi2017icarl} & 58.37\small{±3.51} & 62.49\small{±5.42} & 46.81\small{±0.41} & 52.51\small{±0.44} & 22.53\small{±0.62} & 26.38\small{±0.23} \\
ER~\cite{riemer2019learning} & 44.46\small{±2.87} & 58.84\small{±3.85} & 23.41\small{±1.15} & 40.47\small{±0.95} & 10.13\small{±0.39} & 25.12\small{±0.56}\\
BiC~\cite{wu2019large} & 52.61\small{±5.37} & 71.95\small{±1.82} & 37.82\small{±1.67} & 47.17\small{±1.17} & 15.36\small{±1.31} & 18.67\small{±0.57} \\
LUCIR~\cite{hou2019learning} & 49.18\small{±7.61} & 65.26\small{±2.54} & 37.91\small{±1.18} & 50.42\small{±0.76} & 28.79\small{±0.51} & 31.64\small{±0.51} \\
DER~\cite{buzzega2020dark} & 57.92\small{±1.91} & 68.65\small{±1.82} & 34.83\small{±2.09} & 50.12\small{±0.75} & 15.14\small{±1.29} & 20.35\small{±0.35} \\
DER++~\cite{buzzega2020dark} & 62.19\small{±1.94} & 70.10\small{±1.65} & 37.69\small{±0.97} & 51.82\small{±1.04} & 19.43\small{± 1.63} & 36.89\small{± 1.16} \\
ER-ACE~\cite{caccia2021new} & 62.19\small{±1.67} & 71.15\small{±1.08} & 37.81\small{±0.54} & 49.77\small{±0.34} & 20.42\small{±0.39} & 37.76\small{±0.53} \\
XDER~\cite{boschini2022class} & 64.10\small{±1.08} & 67.42\small{±2.16} & \textcolor{mygreen}{48.14\small{±0.34}} & \textcolor{mygreen}{57.57\small{±0.84} } & 29.12\small{±0.47} & \textcolor{mygreen}{46.12\small{±0.46}} \\
CLS-ER~\cite{arani2022learning} & 64.56\small{±2.63} & 74.27\small{±0.81} & 43.92\small{±0.62} & 54.84\small{±1.30} & \textcolor{mygreen}{30.91\small{±0.59}} & 45.17\small{±0.89}\\
%NPCL~\cite{jha2023npcl} & 63.78\small{±1.70} & 71.34\small{±1.48} & 37.43 & 46.71 & 12.44\small{±0.59} & -\\
SCoMMER~\cite{sarfraz2023sparse} & 66.95\small{±1.52} & 73.64\small{±0.43} & 39.05\small{±0.79} & 49.42\small{±0.85} & 21.47\small{±0.54}  & 37.2\small{±0.70} \\
BFP~\cite{gu2023preserving} & 68.64\small{±2.23} & 73.51\small{±1.54} & 46.70\small{±1.45} & 57.39\small{±0.75} & 28.71\small{±0.55} & 43.17\small{±1.89} \\
SARL~\cite{sarfrazsemantic} & 68.87\small{±1.37} & 73.98\small{±0.46} & 46.69\small{±0.79} & 57.06\small{±0.48} & 28.44\small{±2.30} & 38.83\small{±0.81} \\
%\rowcolor{gray!15} \textbf{Ours (ER+ID)} & 65.50\small{±2.72} & 72.49\small{±1.13} & 42.80\small{±1.57} & 53.67\small{±1.61} & 39.16\small{±0.94} \\
\midrule
\textbf{ER+ID(Ours)} & \textcolor{mygreen}{71.02\small{±1.98}} & 74.74\small{±0.42} & 44.82\small{±0.85} & 56.59\small{±0.35} & 29.88\small{±1.15} & 43.05\small{±1.40} \\
%\rowcolor{gray!15} \textbf{Ours (ER+ID cls-architecture)} & 69.20\small{±1.68} & 75.80\small{±1.36} & 46.32\small{±1.75} & 55.50\small{±0.37} & 44.82\small{±0.33} \\

%\rowcolor{gray!15} \textbf{Ours (BFP+ID)} & 67.46\small{±1.68} & 74.25\small{±1.49} & 47.25\small{±0.52} & 57.83\small{±0.80} & 42.10\small{±0.33} \\
\textbf{BFP+ID (Ours)} & \textcolor{myblue}{71.99\small{±0.98} }& \textcolor{myblue}{76.65\small{±0.63}} & \textcolor{myblue}{48.53\small{±0.95}} & \textcolor{myblue}{57.74\small{±0.64}} & 30.62\small{±0.47} & 43.51\small{±0.59} \\
%\rowcolor{gray!15} \textbf{Ours (BFP+ID cls-architecture)} & 71.46\small{±0.95} & 77.60\small{±0.51} & 47.89\small{±0.49} & 57.72\small{±0.31} & 45.19\small{±0.37} \\
\textbf{CLS-ER+ID (Ours)} & 70.32\small{±1.12} & \textcolor{mygreen}{75.48\small{±0.91}} & 47.44\small{±2.0} & 56.36\small{±0.78} & \textcolor{myblue}{31.62\small{±0.57}} & \textcolor{myblue}{46.17\small{±0.22}}\\
\bottomrule
\end{tabular}%
}
\label{tab:main_results}
\end{table}
\begin{figure}[t]
    \centering
    \begin{minipage}[b]{0.32\textwidth}
        \centering
        \includegraphics[width=\textwidth]{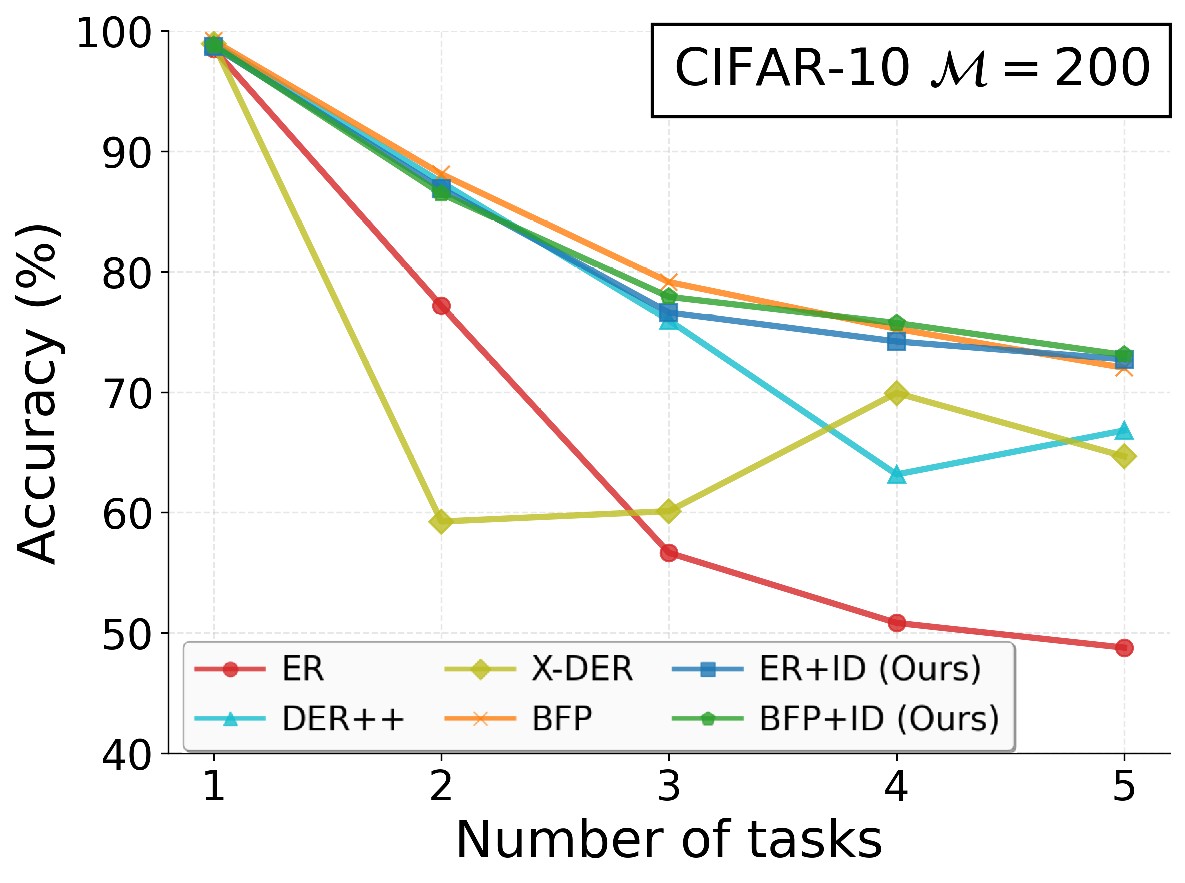}
    \end{minipage}
    \hfill
    \begin{minipage}[b]{0.32\textwidth}
        \centering
        \includegraphics[width=\textwidth]{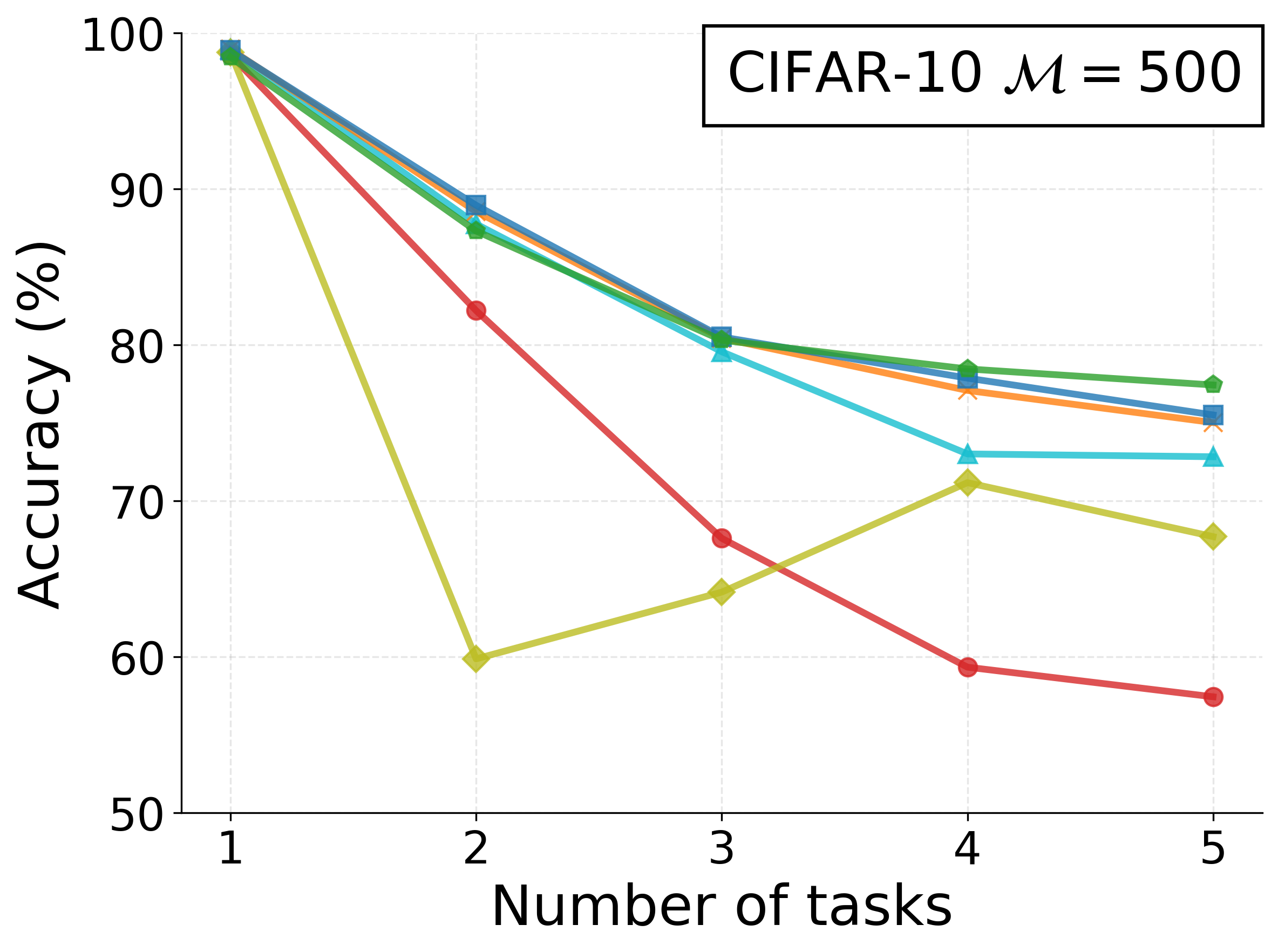}
    \end{minipage}
    \hfill
    \begin{minipage}[b]{0.32\textwidth}
        \centering
        \includegraphics[width=\textwidth]{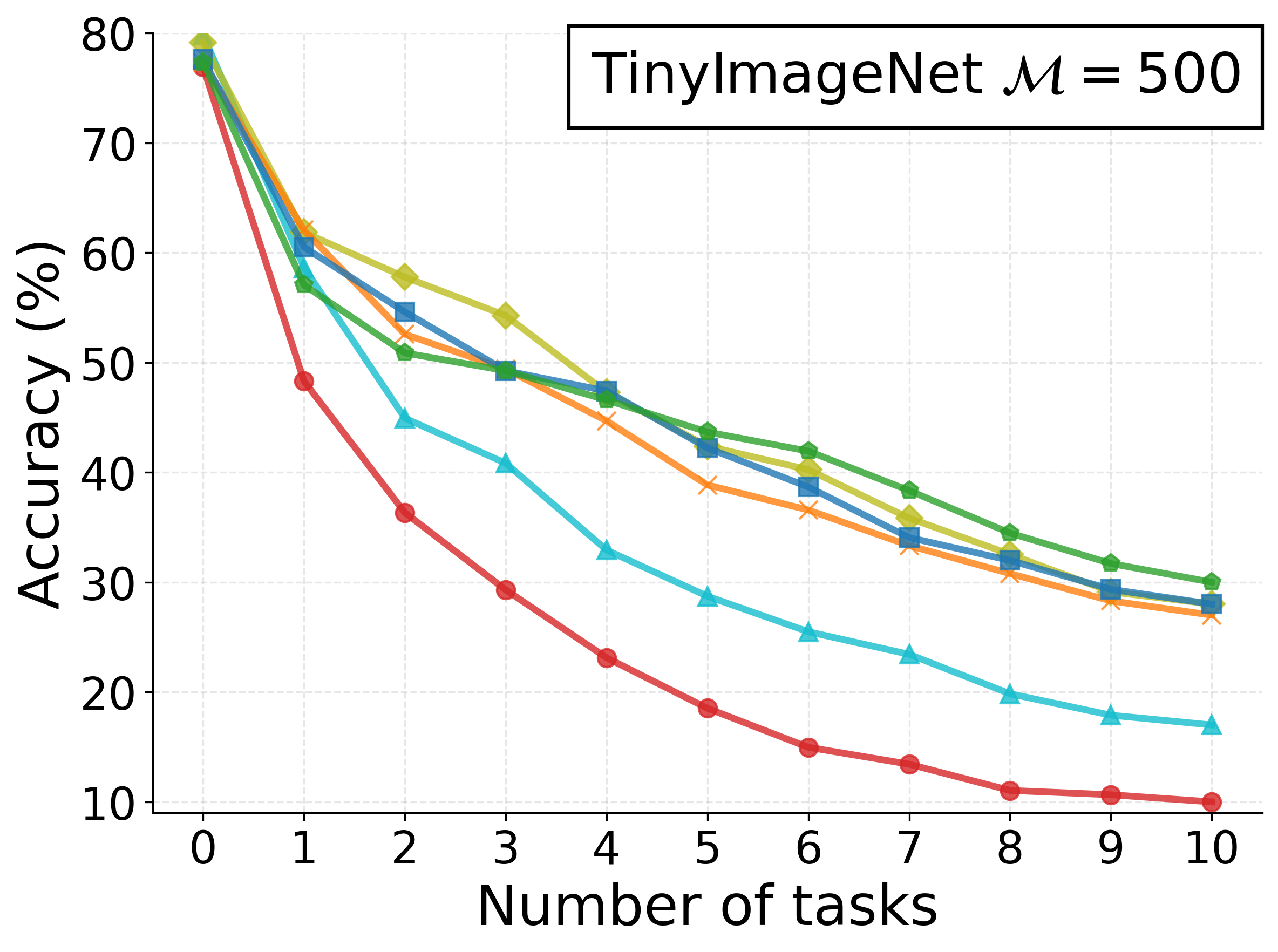}
    \end{minipage}
    \caption{Results on CIFAR-10 and Tiny-ImageNet with different buffer size. It shows the trend of the average test-set accuracy on the observed tasks.}
    \label{fig:results}
\end{figure}

\noindent\textbf{Training Details.}
We adopt the standard experimental protocols following~\citet{gu2023preserving,boschini2022class}. All methods use a ResNet-18 backbone~\cite{he2016deep} trained from scratch with an SGD optimizer. For a fair comparison, we employ uniform settings across all methods (including epochs, batch sizes, and optimizer configurations). Datasets are split as follows: 5 tasks for CIFAR-10, and 10 tasks each for CIFAR-100 and TinyImageNet. We report the average results over 5 independent runs with different random seeds to ensure statistical reliability. Comprehensive hyperparameter settings and further implementation details are provided in the appendix.

%Our method also uses class-balanced reservoir sampling method. 
%The batch size is set to 32 for both the current task samples and the samples from the memory buffer. 
%All experiments are repeated five times with different random seeds to ensure reliability. More details about experimental setting can be found in the appendix.
\begin{comment}
\begin{table}[t]
\centering
\small
\caption{Model calibration ECE errors comparison on different datasets with varying buffer sizes. 
         }
\label{tab:main_results}
\begin{tabular}{lcc}
\toprule
\textbf{Method} & \textbf{S-CIFAR-10 } & \textbf{S-CIFAR-100 } \\
\midrule
DER      & 3.43 & 9.67 \\
ICARL    & 6.94 & 13.82 \\
BFP      & 3.73 & 0.5441* \\
ER+ID (ours)    & \textbf{1.98} & \textbf{0.1995*} \\
\bottomrule
\end{tabular}
\end{table}
\end{comment}

\subsection{Results}
\noindent\textbf{Comparison with the state-of-the-art methods.}
%Across all settings, the NPCL boosts
%the performance of the ER and achieves either comparable or better results against the state-of-theart (SOTA), e.g., DER. 
We evaluate our method against state-of-the-art continual learning approaches across three benchmark datasets with different memory buffer sizes: CIFAR-10, CIFAR-100, and Tiny-ImageNet. The Final Average Accuracies in the class incremental learning setting on different benchmarks are reported in Table~\ref{tab:main_results}.  Our method outperforms all rehearsal-based methods on three datasets. Notably, our method outperforms the second best method BFP by up to 3$\%$ on CIFAR-10, which shows that our method remains highly effective even on a small-scale benchmark. Though outperforming XDER only slightly in FAA on CIFAR-100 and Tiny-ImageNet, our approach attains this accuracy with markedly lower computational cost, which can be shown in Figure~\ref{fig:ab_results} (a). Figure~\ref{fig:results} shows that IDER has better performance at most intermediate tasks and also the final one. In addition, Table~\ref{tab:gcil_results} highlights the advantage of IDER in the challenging GCIL setting, which tests the model’s ability to deal with class imbalance and to continuously integrate knowledge from overlapping classes. The results in such a challenging setting prove the benefits of idempotence, which encourages the model to produce more robust representations to identify concepts clearly. This ability of IDER shows the potential for realistic continual learning.

\noindent\textbf{Plug-and-play with other rehearsal-based methods.}
Considering the effectiveness and simplicity of idempotence, it is natural to consider whether it can be integrated into other rehearsal-based methods. Table~\ref{tab:main_results} shows consistent performance improvements on various datasets with this integration. Enforcing idempotence boosts FAA by a significant margin, especially for ER ( 26$\%$ on CIFAR-10 with buffer size 200 and 21$\%$ on CIFAR-100 with buffer size 500). The results in GCIL in Table~\ref{tab:gcil_results} can also prove that IDER, by enforcing model idempotence, is complementary to other methods in relieving forgetting. It is worth mentioning that in more challenging setting, the performance gains can be obvious in practice. Combined with CLS-ER, in traditional CIL, idempotence yields a gain of about 3.5$\%$ on CIFAR-100 with buffer size 500, while in GCIL, the gains can reach 8$\%$. This additionally demonstrates the potential of this mathematical property to address catastrophic forgetting for even more challenging CL scenarios. 

%Notably, with low buffer size memories, our method  improves the Final Average Accuracy (FAA) by a significant margin. For example, on CIFAR-100 with a buffer size of 500, we achieve a clear improvement by only adding the idempotence loss on ER, improving 21.41\% FAA, while increasing 7.13\% compared with the baseline DER++. It shows that our method could effectively preserve the previous knowledge with limited samples for previous classes.  What's more, the results reveal that combining our method with state-of-the-art method yields the better performance, suggesting that our method can effectively leverage complementary strengths to further mitigate catastrophic forgetting and improve overall continual learning performance. In addition, our method could achieve comparable performance with XDER on complex datasets, yet also having strong performance on the simpler dataset, such as CIFAR-10, with markedly lower computational cost which can be shown in Figure.
\begin{table}[t]
\centering
%\captionsetup{font=small}
\caption{Comparison of Final Average Accuracy (FAA) across different continual learning methods on GCIL-CIFAR-100 dataset. All experiments are repeated 5 times with different seeds. Absolute gains are indicated in \textcolor{mygreen}{green}.}
\renewcommand{\arraystretch}{1.3}
\resizebox{0.85\textwidth}{!}{
\begin{tabular}{l cc cc cc cc}
\toprule
\multirow{2}{*}{\textbf{Method}} & \multicolumn{4}{c}{\textbf{Uniform}} & \multicolumn{4}{c}{\textbf{Longtail}} \\
& Buffer 200 & $\Delta$ & Buffer 500 & $\Delta$ & Buffer 200 & $\Delta$ & Buffer 500 & $\Delta$ \\
\midrule
Joint (upper bound) & \multicolumn{4}{c}{58.36\small{±1.02}}  & \multicolumn{4}{c}{56.94\small{±1.56}}  \\
\midrule

DER++~\cite{buzzega2020dark} & 19.36\small{±0.65} &  & 33.66\small{±0.96} &  & 27.05\small{±1.11} &  & 25.98\small{±0.81} &  \\
SCoMMER~\cite{sarfraz2023sparse} & 28.56\small{±2.26} &  & 35.70\small{±0.86} &  & 28.47\small{±1.12} &  & 32.99\small{±0.49} &  \\
\midrule
ER~\cite{riemer2019learning} & 16.34\small{±0.74} &  & 28.76\small{±0.66} &  & 19.55\small{±0.69} &  & 20.02\small{±1.05} &  \\
\rowcolor{gray!15} \textbf{Ours (ER+ID)} & 26.66\small{±0.63} & \textcolor{mygreen}{\textbf{+10.32}} & 40.54\small{±0.46} & \textcolor{mygreen}{\textbf{+11.78}} & 30.04\small{±0.58} & \textcolor{mygreen}{\textbf{+10.49}} & 35.92\small{±0.35} & \textcolor{mygreen}{\textbf{+15.90}} \\ 
CLS-ER~\cite{arani2022learning} & 22.37\small{±0.48} &  & 36.80\small{±0.34} &  & 28.34\small{±0.99} &  & 28.35\small{±0.72} &  \\
\rowcolor{gray!15} \textbf{Ours (CLS-ER+ID)} & 31.17\small{±1.62} & \textcolor{mygreen}{\textbf{+8.80}} & 37.57\small{±1.81} & \textcolor{mygreen}{\textbf{+0.77}} & 34.08\small{±0.45} & \textcolor{mygreen}{\textbf{+5.74}} & 36.75\small{±0.62} & \textcolor{mygreen}{\textbf{+8.40}} \\
SARL~\cite{sarfrazsemantic} & 36.20\small{±0.46} &  & 38.73\small{±0.66} &  & 34.13\small{±1.07} &  & 34.64\small{±0.49} &  \\
\rowcolor{gray!15} \textbf{Ours (SARL+ID)} & 36.45\small{±0.37} & \textcolor{mygreen}{\textbf{+0.25}} & 39.65\small{±0.43} & \textcolor{mygreen}{\textbf{+0.92}} & 35.04\small{±0.54} & \textcolor{mygreen}{\textbf{+0.91}} & 35.67\small{±0.74} & \textcolor{mygreen}{\textbf{+1.03}} \\
\bottomrule
\end{tabular}%
}
\label{tab:gcil_results}
\end{table}
\noindent\textbf{Idempotence Improves prediction Reliability.}
As previously reported by\citet{guo2017calibration}, DNN are uncalibrated, often tending towards overconfidence. \citet{arani2022learning} show that this problem is pronounced in continual learning where the models  tend to be biased towards recent tasks. Following~\citet{boschini2022class,jha2023npcl} we evaluate the calibration errors for different CL baselines using the well-established Expected Calibration Error (ECE), which is shown in Table~\ref{tab:ece_results}. Table~\ref{tab:ece_results} shows that IDER consistently reduce the calibration error. In general, IDER benefits CL models in confidence calibration which demonstrates the ability of IDER to make reliable predictions. Comparing with post-hoc uncertainty calibration methods for CIL~\cite{hwang2025t,li2024calibration}, IDER achieves comparable performances across different datasets. This strong correlation between improved calibration and higher accuracy suggests that by producing more reliable confidence estimates, the model mitigates overconfidence on its own predictions (potentially incorrect), thereby facilitating a more stable and effective learning process that leads to better overall performance.
\begin{table}[t]
\centering
%\captionsetup{font=small}
\caption{Comparison of Expected Calibration Error (ECE) across different continual learning methods on CIFAR-10 and CIFAR-100 dataset. All experiments are repeated 5 times with different seeds. Results of NPCL are imported from its
original work~\cite{gu2023preserving}.  Absolute improvements (lower ECE) are indicated in \textcolor{myred}{red}.}
\renewcommand{\arraystretch}{1.3}
\resizebox{1\textwidth}{!}{
\begin{tabular}{l cc cc cc cc cc cc}
\toprule
\multirow{2}{*}{\textbf{Method}} & \multicolumn{4}{c}{\textbf{CIFAR-10}} & \multicolumn{4}{c}{\textbf{CIFAR-100}} & \multicolumn{4}{c}{ \textbf{Tiny-ImageNet}} \\
& Buffer 200 & $\Delta$ & Buffer 500 & $\Delta$ & Buffer 500 & $\Delta$ & Buffer 2000 & $\Delta$ &  Buffer 500 &  $\Delta$ &  Buffer 4000 &  $\Delta$\\
\midrule
DER~\cite{buzzega2020dark} & 29.91 &  & 16.20 &  & 24.84 &  & 10.79 &  & 22.80 & &  10.52 &\\
NPCL~\cite{jha2023npcl} & 21.03 &  & - &  & 19.95 &  & - &  & - & & - &\\
RC~\cite{li2024calibration} & 16.39 &  & 12.84 &  & 19.43 &  & 19.31 &  & 21.32 & & 16.49 &\\
T-CIL~\cite{hwang2025t} &  22.50 &  & 10.51 &  & 15.79 &  & 8.67 &  & 14.50 & & 10.30 &\\

\midrule
ER~\cite{riemer2018learning} & 45.53 &  & 32.69 &  & 64.59 &  & 45.64 &  & 67.50 & & 51.37 &\\
\rowcolor{gray!15} \textbf{Ours (ER+ID)} & 12.36 & \textcolor{myred}{\textbf{-33.17}} & 11.73 & \textcolor{myred}{\textbf{-20.96}} & 13.65 & \textcolor{myred}{\textbf{-50.94}} & 12.87 & \textcolor{myred}{\textbf{-32.77}} & 21.55 & \textcolor{myred}{\textbf{-49.45}}  & 11.14  & \textcolor{myred}{\textbf{-40.23}}\\ 
%CLS-ER~\cite{arani2022learning} & 22.37\small{±0.48} &  & 36.80\small{±0.34} &  & 28.34\small{±0.99} &  & 28.35\small{±0.72} &  \\
%\rowcolor{gray!15} \textbf{Ours (CLS-ER+ID)} & 31.17\small{±1.62} & \textcolor{mygreen}{\textbf{+8.80}} & 37.57\small{±1.81} & \textcolor{mygreen}{\textbf{+0.77}} & 34.08\small{±0.45} & \textcolor{mygreen}{\textbf{+5.74}} & 36.75\small{±0.62} & \textcolor{mygreen}{\textbf{+8.40}} \\
BFP~\cite{gu2023preserving} & 9.83 &  & 9.40 &  & 11.93 &  & 9.28  &  & 9.45 & & 8.25 & \\
\rowcolor{gray!15} \textbf{Ours (BFP+ID)} & 9.30 & \textcolor{myred}{\textbf{-0.53}} & 8.63 & \textcolor{myred}{\textbf{-0.77}} & 8.92 & \textcolor{myred}{\textbf{-3.01}} & 8.29 & \textcolor{myred}{\textbf{-0.99}} & 7.77 & \textcolor{myred}{\textbf{-1.68}} & 6.35 & \textcolor{myred}{\textbf{-1.9}}\\
\bottomrule
\end{tabular}%
}
\label{tab:ece_results}
\end{table}
\FloatBarrier
\vspace*{0.1\baselineskip}
\subsection{Additional Analysis}
\noindent\textbf{Backbone Modification.}
To enforce idempotence, we introduce a lightweight architectural modification by dividing the backbone into two parts (see Section 3.2) so that the model can incorporate the second input. We first evaluate whether this structural change alone affects performance on Split CIFAR-100 with a buffer size of 500. As shown in Table~\ref{tab:ablation_structure}, the modified structure performs similarly to the normal backbone, indicating that the architectural change itself does not alter the baseline performance; therefore, the observed gains mainly come from the proposed idempotent loss rather than the modified architecture. We further ablate different partition points using ResNet-18 on Split CIFAR-100 and Tiny-ImageNet. Table~\ref{tab:layer_comparison} shows that shallower splits amplify noise and destabilize training, whereas deeper splits attenuate the second signal and weaken the idempotence effect. Based on these results, we split the backbone at a mid-layer, which yields the most consistent and strongest performance across different datasets overall.
\begin{comment}
\begin{table}[h]
\centering
    \centering
    \renewcommand{\arraystretch}{1.2}
    \captionsetup{font=small}
    \caption{Comparison of the performances with modified backbone and normal backbone. The results are almost same, which shows that the modified structure is reasonable and don't influence the performance.}
    \resizebox{0.6\linewidth}{!}{ 
    \begin{tabular}{lcccc}
        \hline
        \textbf{Model} & \textbf{Method} & \textbf{Accuracy (\%)} & \textbf{Forgetting (\%)} \\
        \hline
        \multirow{2}{*}{Normal ResNet-18} 
            & Finetune & 8.29 & 90.52 \\
            & ER       & 24.36 & 71.30 \\
        \hline
        \multirow{2}{*}{Modified ResNet-18} 
            & Finetune & 8.23 & 90.58 \\
            & ER       & 24.73 & 70.61 \\
        \hline
    \end{tabular}}
\label{tab:ablation_structure}
\end{table}
\end{comment}
\begin{table}[H]
    \centering

    \begin{minipage}[t]{0.48\textwidth}
        \vspace{0pt} % ensure top alignment
        \centering
        \renewcommand{\arraystretch}{1.2}
        \captionsetup{width=\linewidth}

        \caption{Comparison of performances using the normal vs.\ modified backbone. Similar results indicate that the architectural change itself does not affect baseline performance.}
        \label{tab:ablation_structure}

        \resizebox{0.98\linewidth}{!}{%
        \begin{tabular}{lccc}
            \toprule
            \textbf{Model} & \textbf{Method} & \textbf{Accuracy (\%)} & \textbf{Forgetting (\%)} \\
            \midrule
            \multirow{2}{*}{Normal ResNet-18}
                & Finetune & 8.29 & 90.52 \\
                & ER       & 24.36 & 71.30 \\
            \midrule
            \multirow{2}{*}{Modified ResNet-18}
                & Finetune & 8.23 & 90.58 \\
                & ER       & 24.73 & 70.61 \\
            \bottomrule
        \end{tabular}%
        }
    \end{minipage}
    \hfill
    \begin{minipage}[t]{0.48\textwidth}
        \vspace{0pt} % ensure top alignment
        \centering
        \renewcommand{\arraystretch}{1.2}
        \captionsetup{width=\linewidth}

        \caption{Comparisons with different partition points on Split CIFAR-100 and Tiny-ImageNet using ResNet-18. Mid-layer partitioning yields the most stable and  best performance.}
        \label{tab:layer_comparison}

        \resizebox{0.98\linewidth}{!}{%
        \begin{tabular}{lcc}
            \toprule
            \textbf{Layer Selection} & \textbf{CIFAR-100} & \textbf{Tiny-ImageNet} \\
            \midrule
            Shallower (5th layer) & 40.51 & 42.10 \\
            Deeper (13th layer)   & 43.09 & 42.28 \\
            \textbf{Ours (9th layer)} & \textbf{44.82} & \textbf{43.05} \\
            \bottomrule
        \end{tabular}%
        }
    \end{minipage}
\end{table}

\begin{figure}[t]
    \centering
    \begin{minipage}[b]{0.32\textwidth}
        \centering
        \includegraphics[width=\textwidth,height=3cm]{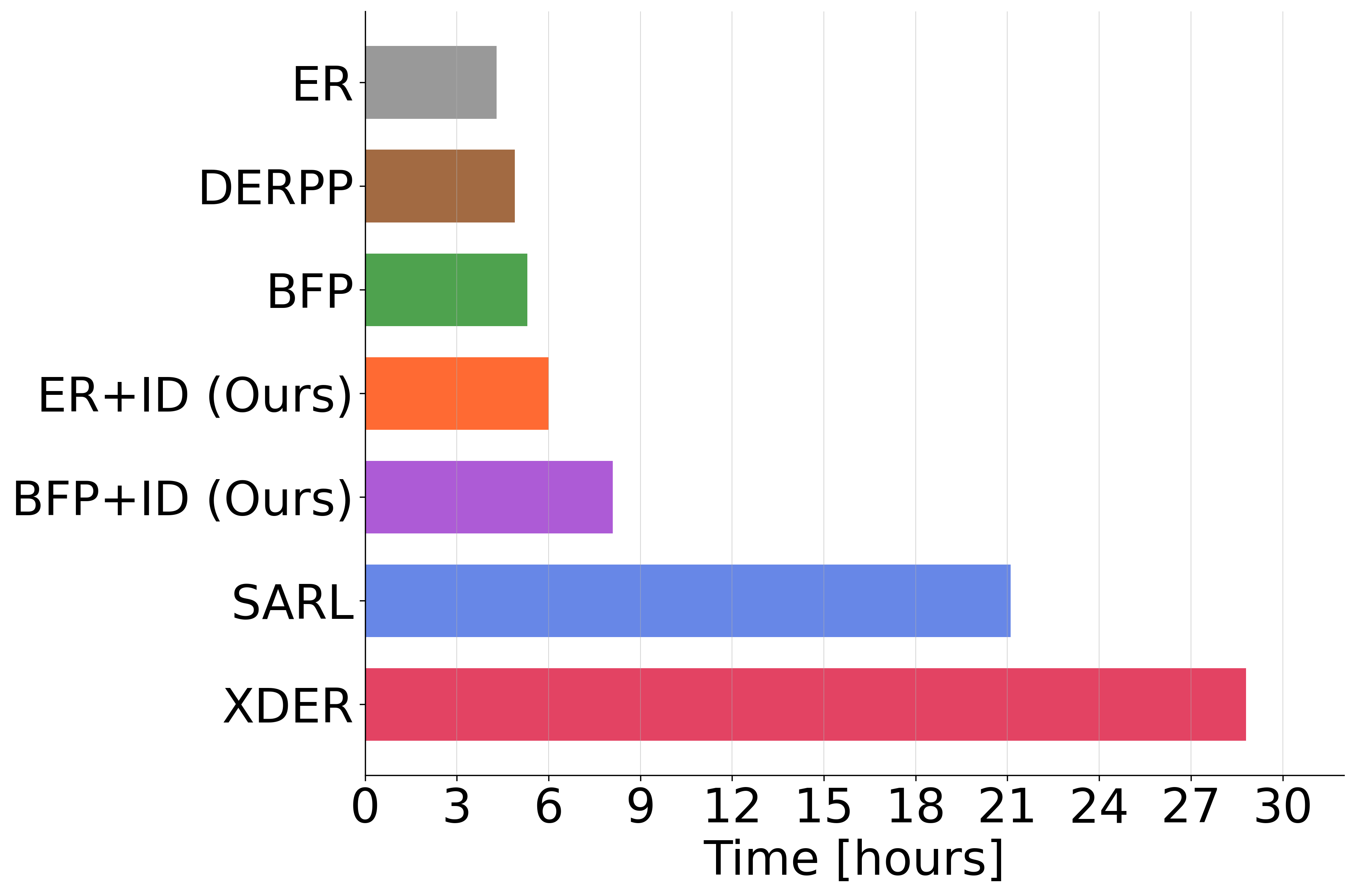}
        \vspace{-0.2cm}
        \centerline{(a) Training Times}
    \end{minipage}
    \hfill
    \begin{minipage}[b]{0.32\textwidth}
        \centering
        \includegraphics[width=\textwidth,height=3cm]{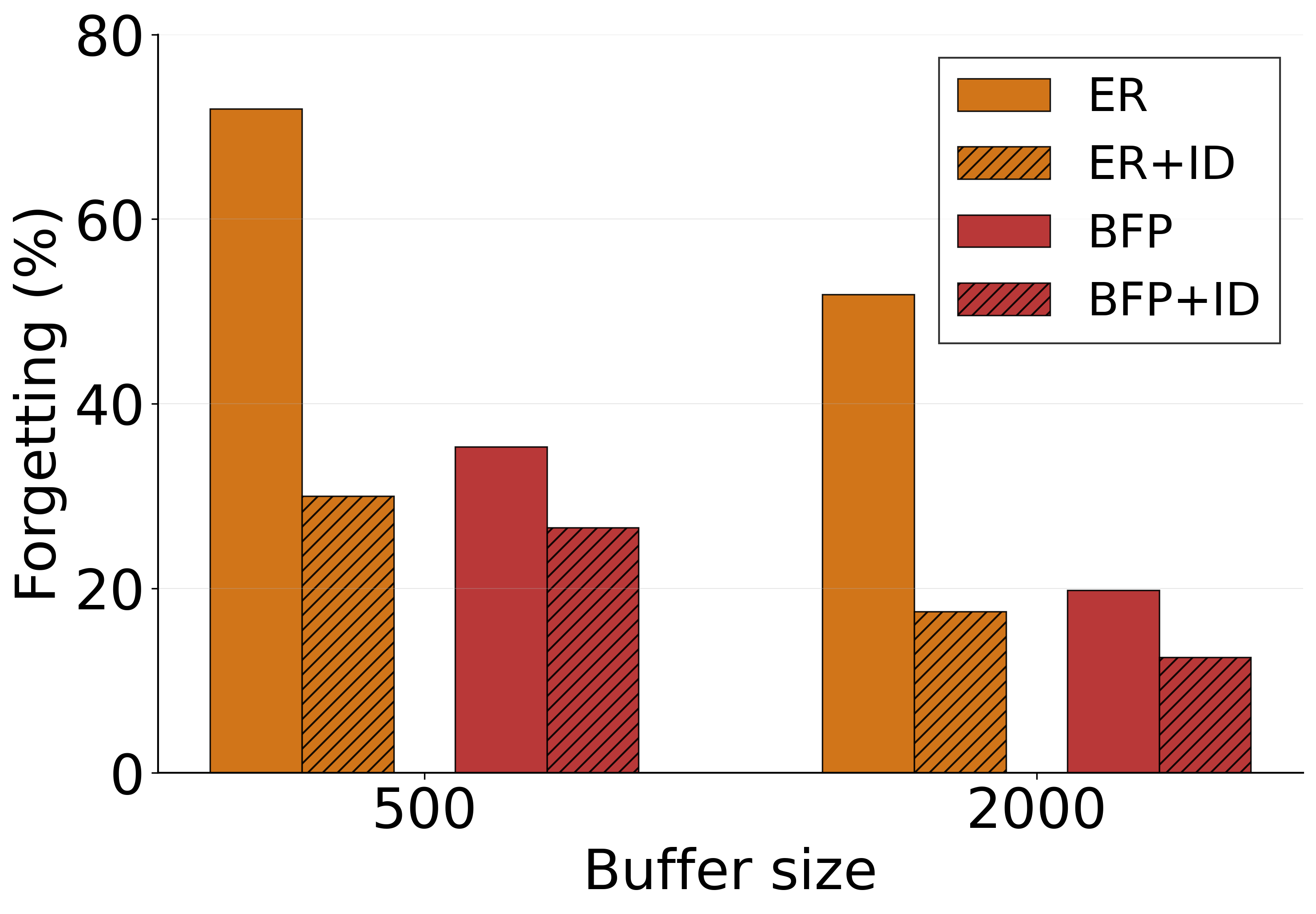}
        \vspace{-0.2cm}
        \centerline{(b) Forgetting on CIFAR-100}
    \end{minipage}
    \hfill
    \begin{minipage}[b]{0.32\textwidth}
        \centering
        \includegraphics[width=\textwidth,height=3cm]{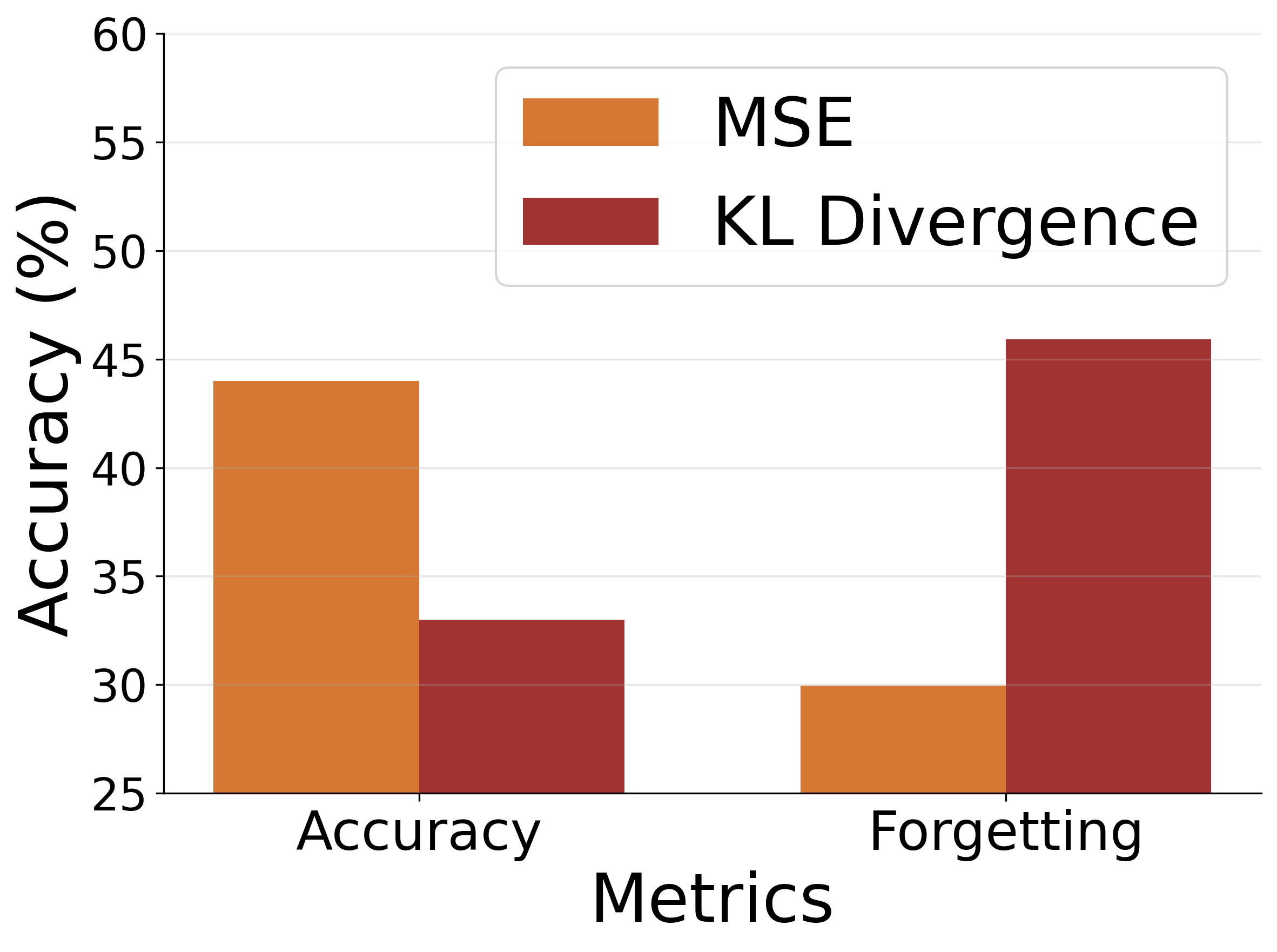}
        \vspace{-0.2cm}
        \centerline{(c) Distance Metrics}
    \end{minipage}
    \caption{Results for model analysis. (a) the training time of different methods on Split TinyImageNet with buffer 500. (b) the Final Forgetting (FF) measures on Split CIFAR-100 with different buffer sizes. (c) the performances on Split CIFAR-100 using different distance metrics for idempotent distillation loss.}
    \label{fig:ab_results}
\end{figure}

\noindent\textbf{Idempotence improves forgetting.}
The Figure~\ref{fig:ab_results} (b) shows Final Forgetting (FF) measured on the Split CIFAR-100 dataset with
different buffer sizes. Our method consistently reduces forgetting, which shows that enforcing idempotence improves accuracy while mitigating the forgetting problem.

\noindent\textbf{On training time.}
Figure~\ref{fig:ab_results} (a) compares the training times of various methods. As expected, our proposed method introduces minimal computational overhead when integrated into existing replay-based methods. In practice, IDER only adds one extra forward pass, whereas X-DER involves additional training components (e.g., additional update steps beyond standard replay), leading to notably higher runtime.This highlights IDER's practicality as a lightweight and effective method.

\noindent\textbf{Comparison with different distance metrics.}
%Table~\ref{tab:loss_ab} reveals the importance of each IDER element in improving performance on 
%split CIFAR-100 with buffer 500. It shows that the importance of enforcing idempotence on buffer data, which improves the stability of the model. 
The figure~\ref{fig:ab_results} (c) shows the effect of different distance metrics for computing the Idempotent distillation loss. While both MSE and KL divergence are well-established metrics for quantifying loss distance, MSE provides better performance. The reason is that MSE avoids the information loss occurring in probability
space due to the squashing function.

\begin{figure}[t]
    \centering
    \includegraphics[width=0.9\linewidth]{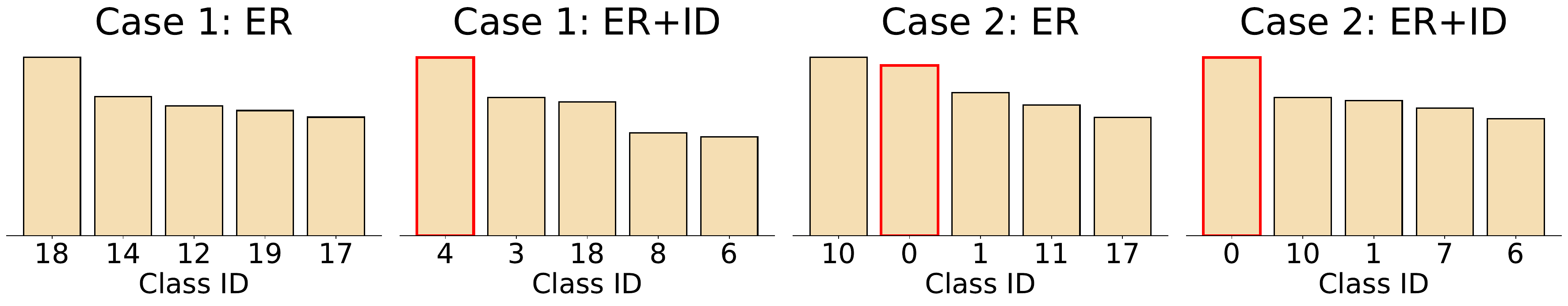}
    %\captionsetup{font={color=blue},labelfont={color=blue}}
    \caption{ Visualizations of the predictions on Split CIFAR-100. It shows top-5 prediction probability produced by ER and ER+ID (ours) after training the first two tasks. The first 10 classes (class ids from 0 to 9) belong to task 1 and the next 10 classes (class ids from 10 to 19) belong to task 2. The ground-truth class is highlighted with red boxes.}
    \label{fig:example_p}
\end{figure}

\noindent\textbf{Visualizations of concrete example.}  To explore how enforcing idempotence mitigates recency bias and improves calibration,  we visualize predictions for test examples using ER and ER+ID on Split CIFAR-100 with a buffer size of 500 and 10 incremental tasks.  Figure~\ref{fig:example_p} shows top-5 predictions after learning the first two tasks on Split CIFAR-100. ER exhibits clear recency bias: classes from the current task (class ids from 10 to 19) receive inflated scores, leading to misclassifications. Integrated with idempotence loss, the predictions get corrected as true class is promoted to top-1 prediction and overconfidence on new classes notably reduced.

\noindent\textbf{Cross-Platform Performance.}  To assess the hardware sensitivity of our results, we run ER+ID with the same configuration on two different platforms: NVIDIA RTX 4090 and Huawei Ascend 910B. As Table~\ref{tab:example_a} shows, the performance is consistent across platforms: the differences are small suggesting that the gains from IDER are not hardware-specific. Minor discrepancies are expected due to differences in kernels, numerical precision, and non-determinism in deep learning libraries.

\begin{table}[H]
\centering
%\captionsetup{font=small}
\caption{Performance comparison for ER+ID between NVIDIA RTX 4090 and Huawei Ascend
910B. All experiments are repeated 5 times with different seeds.}
\label{tab:example_a}
\renewcommand{\arraystretch}{1.3}
\resizebox{0.8\textwidth}{!}{
\begin{tabular}{l cc cc c}
\toprule
\multirow{2}{*}{\textbf{Platform}} &
\multicolumn{2}{c}{\textbf{CIFAR-10}} &
\multicolumn{2}{c}{\textbf{CIFAR-100}} &
\multicolumn{1}{c}{\textbf{Tiny-ImageNet}} \\
& Buffer 200 & Buffer 500 & Buffer 500 & Buffer 2000 & Buffer 4000 \\
\midrule
RTX 4090 & 71.02\small{±1.98} & 74.74\small{±0.42}  & 44.82\small{±0.85} & 56.59\small{±0.35} & 43.05\small{±1.40} \\
Ascend 910B  & 71.13\small{±0.92} & 74.68\small{±0.37} & 44.87\small{±0.76} & 56.44\small{±0.41} & 43.01\small{±1.26} \\
\bottomrule
\end{tabular}
}
\end{table}

\section{Conclusion}

In this paper, we propose Idempotent Experience Replay (IDER), a simple and effective method designed to mitigate catastrophic forgetting and improve predictive reliability in continual learning. Our approach adapts the training loss and introduces idempotence distillation loss for CL methods to encourage . Extensive experiments demonstrate that IDER consistently improves performance across multiple datasets and diverse continual learning settings. Our results show that enforcing idempotence enables a balance between stability and plasticity while yielding better calibrated predictions. Our method, requiring only two forward passes without additional parameters and seamlessly integrated with other CL approaches, shows promise for deployment of CL models in real-world scenarios. We hope this work inspires future research to place greater emphasis on uncertainty-aware continual learning.
%and to develop more effective method to build trustworthy continual learning in real-world applications. 
We also plan to explore the potential of idempotence property in different domains.
%not only within existing applications like out-of-distribution detection and generative network, but also in new domains in the future.

\newpage
\section*{ACKNOWLEDGEMENT}
This project is supported by the National Natural Science Foundation of China (No.\ 62406192), the Shanghai Municipal Special Program for Basic Research on General AI Foundation Models (Grant No.\ 2025SHZDZX025G03), Opening Project of the State Key Laboratory of General Artificial Intelligence (No.\ SKLAGI2024OP12), the Tencent WeChat Rhino-Bird Focused Research Program, Kuaishou Technology, and the SJTU Kunpeng \& Ascend Center of Excellence.

{\small
\bibliography{ICLR2026/reference}

\begin{thebibliography}{41}
\providecommand{\natexlab}[1]{#1}
\providecommand{\url}[1]{\texttt{#1}}
\expandafter\ifx\csname urlstyle\endcsname\relax
  \providecommand{\doi}[1]{doi: #1}\else
  \providecommand{\doi}{doi: \begingroup \urlstyle{rm}\Url}\fi

\bibitem[Arani et~al.(2022)Arani, Sarfraz, and Zonooz]{arani2022learning}
Elahe Arani, Fahad Sarfraz, and Bahram Zonooz.
\newblock Learning fast, learning slow: A general continual learning method based on complementary learning system.
\newblock \emph{arXiv preprint arXiv:2201.12604}, 2022.

\bibitem[Boschini et~al.(2022)Boschini, Bonicelli, Buzzega, Porrello, and Calderara]{boschini2022class}
Matteo Boschini, Lorenzo Bonicelli, Pietro Buzzega, Angelo Porrello, and Simone Calderara.
\newblock Class-incremental continual learning into the extended der-verse.
\newblock \emph{IEEE transactions on pattern analysis and machine intelligence}, 45\penalty0 (5):\penalty0 5497--5512, 2022.

\bibitem[Buzzega et~al.(2020)Buzzega, Boschini, Porrello, Abati, and Calderara]{buzzega2020dark}
Pietro Buzzega, Matteo Boschini, Angelo Porrello, Davide Abati, and Simone Calderara.
\newblock Dark experience for general continual learning: a strong, simple baseline.
\newblock \emph{Advances in neural information processing systems}, 33:\penalty0 15920--15930, 2020.

\bibitem[Buzzega et~al.(2021)Buzzega, Boschini, Porrello, and Calderara]{buzzega2021rethinking}
Pietro Buzzega, Matteo Boschini, Angelo Porrello, and Simone Calderara.
\newblock Rethinking experience replay: a bag of tricks for continual learning.
\newblock In \emph{2020 25th International Conference on Pattern Recognition (ICPR)}, pp.\  2180--2187. IEEE, 2021.

\bibitem[Caccia et~al.(2021)Caccia, Aljundi, Asadi, Tuytelaars, Pineau, and Belilovsky]{caccia2021new}
Lucas Caccia, Rahaf Aljundi, Nader Asadi, Tinne Tuytelaars, Joelle Pineau, and Eugene Belilovsky.
\newblock New insights on reducing abrupt representation change in online continual learning.
\newblock \emph{arXiv preprint arXiv:2104.05025}, 2021.

\bibitem[Chaudhry et~al.(2019)Chaudhry, Rohrbach, Elhoseiny, Ajanthan, Dokania, Torr, and Ranzato]{chaudhry2019continual}
Arslan Chaudhry, Marcus Rohrbach, Mohamed Elhoseiny, Thalaiyasingam Ajanthan, P~Dokania, P~Torr, and M~Ranzato.
\newblock Continual learning with tiny episodic memories.
\newblock In \emph{Workshop on Multi-Task and Lifelong Reinforcement Learning}, 2019.

\bibitem[Durasov et~al.(2024{\natexlab{a}})Durasov, Dorndorf, Le, and Fua]{durasov2024zigzag}
Nikita Durasov, Nik Dorndorf, Hieu Le, and Pascal Fua.
\newblock Zigzag: Universal sampling-free uncertainty estimation through two-step inference.
\newblock \emph{Transactions on Machine Learning Research}, 2024{\natexlab{a}}.
\newblock ISSN 2835-8856.

\bibitem[Durasov et~al.(2024{\natexlab{b}})Durasov, Oner, Donier, Le, and Fua]{durasov2024enabling}
Nikita Durasov, Doruk Oner, Jonathan Donier, Hieu Le, and Pascal Fua.
\newblock Enabling uncertainty estimation in iterative neural networks.
\newblock In \emph{Forty-first International Conference on Machine Learning}, 2024{\natexlab{b}}.

\bibitem[Durasov et~al.(2024{\natexlab{c}})Durasov, Shocher, Oner, Chechik, Efros, and Fua]{durasov20243}
Nikita Durasov, Assaf Shocher, Doruk Oner, Gal Chechik, Alexei~A Efros, and Pascal Fua.
\newblock It3: Idempotent test-time training.
\newblock \emph{arXiv preprint arXiv:2410.04201}, 2024{\natexlab{c}}.

\bibitem[Farajtabar et~al.(2020)Farajtabar, Azizan, Mott, and Li]{farajtabar2020orthogonal}
Mehrdad Farajtabar, Navid Azizan, Alex Mott, and Ang Li.
\newblock Orthogonal gradient descent for continual learning.
\newblock In \emph{International conference on artificial intelligence and statistics}, pp.\  3762--3773. PMLR, 2020.

\bibitem[Gu et~al.(2023)Gu, Shim, and Shkurti]{gu2023preserving}
Qiao Gu, Dongsub Shim, and Florian Shkurti.
\newblock Preserving linear separability in continual learning by backward feature projection.
\newblock In \emph{Proceedings of the IEEE/CVF Conference on Computer Vision and Pattern Recognition}, pp.\  24286--24295, 2023.

\bibitem[Guo et~al.(2017)Guo, Pleiss, Sun, and Weinberger]{guo2017calibration}
Chuan Guo, Geoff Pleiss, Yu~Sun, and Kilian~Q Weinberger.
\newblock On calibration of modern neural networks.
\newblock In \emph{International conference on machine learning}, pp.\  1321--1330. PMLR, 2017.

\bibitem[He et~al.(2016)He, Zhang, Ren, and Sun]{he2016deep}
Kaiming He, Xiangyu Zhang, Shaoqing Ren, and Jian Sun.
\newblock Deep residual learning for image recognition.
\newblock In \emph{Proceedings of the IEEE conference on computer vision and pattern recognition}, pp.\  770--778, 2016.

\bibitem[Hou et~al.(2019)Hou, Pan, Loy, Wang, and Lin]{hou2019learning}
Saihui Hou, Xinyu Pan, Chen~Change Loy, Zilei Wang, and Dahua Lin.
\newblock Learning a unified classifier incrementally via rebalancing.
\newblock In \emph{Proceedings of the IEEE/CVF conference on computer vision and pattern recognition}, pp.\  831--839, 2019.

\bibitem[Hu et~al.(2025{\natexlab{a}})Hu, Huang, Shen, Yang, Hu, Tang, Chen, Sun, Chang, and Tao]{hu2025tackling}
Jifeng Hu, Sili Huang, Li~Shen, Zhejian Yang, Shengchao Hu, Shisong Tang, Hechang Chen, Lichao Sun, Yi~Chang, and Dacheng Tao.
\newblock Tackling continual offline rl through selective weights activation on aligned spaces.
\newblock In \emph{The Thirty-ninth Annual Conference on Neural Information Processing Systems}, 2025{\natexlab{a}}.

\bibitem[Hu et~al.(2025{\natexlab{b}})Hu, Shen, Huang, Yang, Chen, Sun, Chang, and Tao]{hu2025continual}
Jifeng Hu, Li~Shen, Sili Huang, Zhejian Yang, Hechang Chen, Lichao Sun, Yi~Chang, and Dacheng Tao.
\newblock Continual diffuser (cod): Mastering continual offline rl with experience rehearsal.
\newblock \emph{IEEE Transactions on Neural Networks and Learning Systems}, 2025{\natexlab{b}}.

\bibitem[Hwang et~al.(2025)Hwang, Kim, and Whang]{hwang2025t}
Seong-Hyeon Hwang, Minsu Kim, and Steven~Euijong Whang.
\newblock T-cil: Temperature scaling using adversarial perturbation for calibration in class-incremental learning.
\newblock In \emph{Proceedings of the Computer Vision and Pattern Recognition Conference}, pp.\  15339--15348, 2025.

\bibitem[Jha et~al.(2023)Jha, Gong, Zhao, and Yao]{jha2023npcl}
Saurav Jha, Dong Gong, He~Zhao, and Lina Yao.
\newblock Npcl: Neural processes for uncertainty-aware continual learning.
\newblock \emph{Advances in Neural Information Processing Systems}, 36:\penalty0 34329--34353, 2023.

\bibitem[Jha et~al.(2024)Jha, Gong, and Yao]{jha2024clap4clip}
Saurav Jha, Dong Gong, and Lina Yao.
\newblock Clap4clip: Continual learning with probabilistic finetuning for vision-language models.
\newblock \emph{Advances in neural information processing systems}, 37:\penalty0 129146--129186, 2024.

\bibitem[Kirkpatrick et~al.(2017)Kirkpatrick, Pascanu, Rabinowitz, Veness, Desjardins, Rusu, Milan, Quan, Ramalho, Grabska-Barwinska, et~al.]{kirkpatrick2017overcoming}
James Kirkpatrick, Razvan Pascanu, Neil Rabinowitz, Joel Veness, Guillaume Desjardins, Andrei~A Rusu, Kieran Milan, John Quan, Tiago Ramalho, Agnieszka Grabska-Barwinska, et~al.
\newblock Overcoming catastrophic forgetting in neural networks.
\newblock \emph{Proceedings of the national academy of sciences}, 114\penalty0 (13):\penalty0 3521--3526, 2017.

\bibitem[Le \& Yang(2015)Le and Yang]{le2015tiny}
Yann Le and Xuan Yang.
\newblock Tiny imagenet visual recognition challenge.
\newblock \emph{CS 231N}, 7\penalty0 (7):\penalty0 3, 2015.

\bibitem[LeCun(2022)]{lecun2022path}
Yann LeCun.
\newblock A path towards autonomous machine intelligence version 0.9. 2, 2022-06-27.
\newblock \emph{Open Review}, 62\penalty0 (1):\penalty0 1--62, 2022.

\bibitem[Li et~al.(2024{\natexlab{a}})Li, Piccoli, Cossu, Bacciu, and Lomonaco]{li2024calibration}
Lanpei Li, Elia Piccoli, Andrea Cossu, Davide Bacciu, and Vincenzo Lomonaco.
\newblock Calibration of continual learning models.
\newblock In \emph{Proceedings of the IEEE/CVF Conference on Computer Vision and Pattern Recognition}, pp.\  4160--4169, 2024{\natexlab{a}}.

\bibitem[Li et~al.(2024{\natexlab{b}})Li, Chen, Yu, Chen, and Shen]{li2024sure}
Yuting Li, Yingyi Chen, Xuanlong Yu, Dexiong Chen, and Xi~Shen.
\newblock Sure: Survey recipes for building reliable and robust deep networks.
\newblock In \emph{Proceedings of the IEEE/CVF Conference on Computer Vision and Pattern Recognition}, pp.\  17500--17510, 2024{\natexlab{b}}.

\bibitem[McCloskey \& Cohen(1989)McCloskey and Cohen]{mccloskey1989catastrophic}
Michael McCloskey and Neal~J Cohen.
\newblock Catastrophic interference in connectionist networks: The sequential learning problem.
\newblock In \emph{Psychology of learning and motivation}, volume~24, pp.\  109--165. Elsevier, 1989.

\bibitem[Mi et~al.(2020)Mi, Kong, Lin, Yu, and Faltings]{mi2020generalized}
Fei Mi, Lingjing Kong, Tao Lin, Kaicheng Yu, and Boi Faltings.
\newblock Generalized class incremental learning.
\newblock In \emph{Proceedings of the IEEE/CVF conference on computer vision and pattern recognition workshops}, pp.\  240--241, 2020.

\bibitem[Rebuffi et~al.(2017)Rebuffi, Kolesnikov, Sperl, and Lampert]{rebuffi2017icarl}
Sylvestre-Alvise Rebuffi, Alexander Kolesnikov, Georg Sperl, and Christoph~H Lampert.
\newblock icarl: Incremental classifier and representation learning.
\newblock In \emph{Proceedings of the IEEE conference on Computer Vision and Pattern Recognition}, pp.\  2001--2010, 2017.

\bibitem[Riemer et~al.(2018)Riemer, Cases, Ajemian, Liu, Rish, Tu, and Tesauro]{riemer2018learning}
Matthew Riemer, Ignacio Cases, Robert Ajemian, Miao Liu, Irina Rish, Yuhai Tu, and Gerald Tesauro.
\newblock Learning to learn without forgetting by maximizing transfer and minimizing interference.
\newblock \emph{arXiv preprint arXiv:1810.11910}, 2018.

\bibitem[Riemer et~al.(2019)Riemer, Cases, Ajemian, Liu, Rish, Tu, and Tesauro]{riemer2019learning}
Matthew Riemer, Ignacio Cases, Robert Ajemian, Miao Liu, Irina Rish, Yuhai Tu, and Gerald Tesauro.
\newblock Learning to learn without forgetting by maximizing transfer and minimizing interference.
\newblock In \emph{ICLR (Poster)}, 2019.

\bibitem[Rusu et~al.(2016)Rusu, Rabinowitz, Desjardins, Soyer, Kirkpatrick, Kavukcuoglu, Pascanu, and Hadsell]{rusu2016progressive}
Andrei~A Rusu, Neil~C Rabinowitz, Guillaume Desjardins, Hubert Soyer, James Kirkpatrick, Koray Kavukcuoglu, Razvan Pascanu, and Raia Hadsell.
\newblock Progressive neural networks.
\newblock \emph{arXiv preprint arXiv:1606.04671}, 2016.

\bibitem[Sarfraz et~al.(2023)Sarfraz, Arani, and Zonooz]{sarfraz2023sparse}
Fahad Sarfraz, Elahe Arani, and Bahram Zonooz.
\newblock Sparse coding in a dual memory system for lifelong learning.
\newblock In \emph{Proceedings of the AAAI Conference on Artificial Intelligence}, volume~37, pp.\  9714--9722, 2023.

\bibitem[Sarfraz et~al.(2025)Sarfraz, Arani, and Zonooz]{sarfrazsemantic}
Fahad Sarfraz, Elahe Arani, and Bahram Zonooz.
\newblock Semantic aware representation learning for lifelong learning.
\newblock In \emph{The Thirteenth International Conference on Learning Representations}, 2025.

\bibitem[Shocher et~al.(2023)Shocher, Dravid, Gandelsman, Mosseri, Rubinstein, and Efros]{shocher2023idempotent}
Assaf Shocher, Amil Dravid, Yossi Gandelsman, Inbar Mosseri, Michael Rubinstein, and Alexei~A Efros.
\newblock Idempotent generative network.
\newblock \emph{arXiv preprint arXiv:2311.01462}, 2023.

\bibitem[Vitter(1985)]{vitter1985random}
Jeffrey~S Vitter.
\newblock Random sampling with a reservoir.
\newblock \emph{ACM Transactions on Mathematical Software (TOMS)}, 11\penalty0 (1):\penalty0 37--57, 1985.

\bibitem[Wang et~al.(2024)Wang, Zhang, Su, and Zhu]{wang2024comprehensive}
Liyuan Wang, Xingxing Zhang, Hang Su, and Jun Zhu.
\newblock A comprehensive survey of continual learning: Theory, method and application.
\newblock \emph{IEEE Transactions on Pattern Analysis and Machine Intelligence}, 2024.

\bibitem[Wang et~al.(2022)Wang, Zhang, Lee, Zhang, Sun, Ren, Su, Perot, Dy, and Pfister]{wang2022learning}
Zifeng Wang, Zizhao Zhang, Chen-Yu Lee, Han Zhang, Ruoxi Sun, Xiaoqi Ren, Guolong Su, Vincent Perot, Jennifer Dy, and Tomas Pfister.
\newblock Learning to prompt for continual learning.
\newblock In \emph{Proceedings of the IEEE/CVF conference on computer vision and pattern recognition}, pp.\  139--149, 2022.

\bibitem[Wei et~al.()Wei, Li, Wang, Wang, Kong, Huang, and Sun]{weifirst}
Lai Wei, Yuting Li, Chen Wang, Yue Wang, Linghe Kong, Weiran Huang, and Lichao Sun.
\newblock First sft, second rl, third upt: Continual improving multi-modal llm reasoning via unsupervised post-training.
\newblock In \emph{The Thirty-ninth Annual Conference on Neural Information Processing Systems}.

\bibitem[Wu et~al.(2019)Wu, Chen, Wang, Ye, Liu, Guo, and Fu]{wu2019large}
Yue Wu, Yinpeng Chen, Lijuan Wang, Yuancheng Ye, Zicheng Liu, Yandong Guo, and Yun Fu.
\newblock Large scale incremental learning.
\newblock In \emph{Proceedings of the IEEE/CVF conference on computer vision and pattern recognition}, pp.\  374--382, 2019.

\bibitem[Xu et~al.(2025)Xu, Jin, Wu, Li, Song, Sun, and Yuan]{xu2025llava}
Guowei Xu, Peng Jin, Ziang Wu, Hao Li, Yibing Song, Lichao Sun, and Li~Yuan.
\newblock Llava-cot: Let vision language models reason step-by-step.
\newblock In \emph{Proceedings of the IEEE/CVF International Conference on Computer Vision}, pp.\  2087--2098, 2025.

\bibitem[Yan et~al.(2026)Yan, Song, Fang, Ji, Li, Li, and Sun]{yan2026livemedbench}
Zhiling Yan, Dingjie Song, Zhe Fang, Yisheng Ji, Xiang Li, Quanzheng Li, and Lichao Sun.
\newblock Livemedbench: A contamination-free medical benchmark for llms with automated rubric evaluation.
\newblock \emph{arXiv preprint arXiv:2602.10367}, 2026.

\bibitem[Yao et~al.(2024)Yao, Li, Dai, Xu, Hu, Zhao, and Sun]{yao2024variational}
Dezhong Yao, Sanmu Li, Yutong Dai, Zhiqiang Xu, Shengshan Hu, Peilin Zhao, and Lichao Sun.
\newblock Variational bayes for federated continual learning.
\newblock \emph{arXiv preprint arXiv:2405.14291}, 2024.

\end{thebibliography}
\bibliographystyle{ICLR2026/iclr2026_conference}}

%%%%%%%%%%%%%%%%%%%%%%%%%%%%%%%%%%%%%%%%%%%%%%%%%%%%%%%%%%%%%%%%%%%%%%%%%%%%%%%
%%%%%%%%%%%%%%%%%%%%%%%%%%%%%%%%%%%%%%%%%%%%%%%%%%%%%%%%%%%%%%%%%%%%%%%%%%%%%%%
% APPENDIX
%%%%%%%%%%%%%%%%%%%%%%%%%%%%%%%%%%%%%%%%%%%%%%%%%%%%%%%%%%%%%%%%%%%%%%%%%%%%%%%
%%%%%%%%%%%%%%%%%%%%%%%%%%%%%%%%%%%%%%%%%%%%%%%%%%%%%%%%%%%%%%%%%%%%%%%%%%%%%%%

\clearpage
\appendix
% \onecolumn
\begin{center}
    \Large \textbf{Appendix}\\[0.5cm]
\end{center}

\section{ Extended Discussion: Related Idempotence to Continual learning}

This section presents the theoretical motivation and formulation of the idempotence loss in continual learning by revisiting prior applications of idempotence and elaborating how the proposed loss improves prediction calibration and mitigates catastrophic forgetting

\noindent\textbf{Uncertainty Measurement}

~\citet{durasov2024zigzag} first train the model to satisfy $f(x, 0) \approx y$ and $f(x, y) \approx y$ for each pair $(x,y)$ by minimizing the following loss: 
\begin{equation}
L_{\text{train}} = \left\lVert f(x, 0) - y \right\rVert + \left\lVert f(x, y) - y \right\rVert. 
\end{equation}
Then they define the uncertainty loss as:
\begin{equation}
L(x) = \left\lVert y_1 - y_0 \right\rVert,
\end{equation}
where the network is applied recursively: $y_0 = f(x, 0)$, $y_1 = f(x, y_0)$.

The rationale is as follows:
\begin{enumerate}
  \item If $x$ is in-distribution, then $y_0 \approx y$, and since the network is trained so that $f(x, y) \approx y$, they have $y_1 \approx y_0$. Therefore, the loss is small.
  \item If $x$ is OOD, then $y_0$ is unlikely to approximate the true label. In this case, the pair $(x, y_0)$ is not a valid input as pretraining, leading $y_1$ to be unpredictable and significantly different from $y_0$, resulting in a large $\mathcal{L}(x)$.
\end{enumerate}
Thus, the magnitude of $\lVert y_1 - y_0 \rVert$ serves as a proxy for prediction certainty.

\noindent\textbf{Introducing Idempotence in Continual Learning}

In CL, models are often poorly calibrated and over-confident, a problem exacerbated by recency bias toward new tasks. To mitigate this issue, we require the model to maintain stable predictions on data from previous tasks even after parameter updates induced by new knowledge, as self-consistency indicates that the network’s output is aligned with the learned in-distribution manifold and can make reliable (well-calibrated) prediction. As before, this condition can be translated into enforcing idempotence. We can formalize the desired idempotence condition as $f_t\!\left(x, f_t(x, 0)\right) = f_t(x, 0)$, where $x$ is from both previous and current tasks and $f_t$ represents the current model. In practice, as data from previous tasks can't be obtained, the loss can be defined as: 
\begin{equation}
\mathcal{L} = \sum_{(x,y)\in \mathcal{T}_{t,M}} \left\lVert f_t(x, 0) - f_t\!\left(x, f_t(x, 0)\right) \right\rVert_2^2,
\end{equation}
where $M$ is the buffer memory which stores data from previous tasks.

Minimizing this loss drives the network toward the condition that repeated application of $f_t(x, \cdot)$ does not change the output, which is needed for model to make reliable predictions in CL.

However, minimizing the idempotence loss in CL is not trivial. First, we propose $\mathcal{L}_{ice}$ to train model idempotent for sequential tasks. Second, We modify the idempotence distillation loss by using the model checkpoint at the end of the last task for the second application, which can be rewritten as 
\begin{equation}
\mathcal{L}_{\mathrm{ide}} = \sum_{(x,y)\in \mathcal{T}_{t,M}} \left\lVert f_t(x, 0) - f_{t-1}\!\left(x, f_t(x, 0)\right) \right\rVert_2^2.
\end{equation}
The modification has two benefits:
\begin{itemize}
  \item It prevents training collapse and bias error amplification. Consistent with ~\citet{shocher2023idempotent} and ~\citet{durasov20243} , directly optimizing the idempotence loss induces two gradient pathways:
  1. A desirable pathway that updates $f_{t}(x, 0)$ toward the correct in-distribution manifold.
  2. An undesirable pathway that may cause the manifold to expand, thereby including an incorrect $f_{t}(x, 0)$. For example, if $y_0 = f_{t}(x, 0)$ is an incorrect prediction, then minimizing $\lVert y_0 - y_1 \rVert$ may cause $y_1 = f_{t}(x, y_0)$ to be pulled toward the incorrect $y_0$ and expand the manifold following the wrong gradient pathways, thereby magnifying the error. Another potential problem is to encourage $f_{t}(x, ·)$ to become the identity function, which is trivially idempotent and may cause training collapse. To counteract the latter gradient pathways, a frozen copy of the network is often used. 
  \item It is designed for enforcing idempotence in CL and can serve as a distillation loss. According to eq~\ref{eq:ice loss}, under empirical risk minimization, we can assume that:
  \begin{equation}
  f_t(x, f_t(x, \mathbf{0})) = f_t(x, \mathbf{0}).
  \end{equation}
  Thus, we rewrite the $\mathcal{L}_{ide}$ as:
  \begin{equation}
  \mathcal{L}_{\mathrm{ide}} = \sum_{(x,y)\in \mathcal{T}_{t,M}} \left\lVert f_t(x,f_t(x, 0)) - f_{t-1}\!\left(x, f_t(x, 0)\right) \right\rVert_2^2.
  \label{eq:ide lossv2}
  \end{equation}
  First, according to the same input for $f_t$ and $f_{t-1}$ in eq~\ref{eq:ide lossv2}, this idempotent distillation loss could serve as a standard regularization loss :
  \begin{equation}
    \mathcal{L}_{re} = \sum_{(x,y)\in \mathcal{T}_{t,M}} \left\lVert f_t(x) - f_{t-1}\!\left(x\right) \right\rVert_2^2.
  \end{equation}
  which is often used in CL methods~\cite{gu2023preserving,sarfrazsemantic} to mitigate catastrophic forgetting.
  
  Second, when incorporating the second input that conveys logits from $f_t$, the loss steers the current model $f_t$ to update in a direction where the predictions remain correctly interpretable by the previous model $f_{t-1}$. Consequently, in sequential tasks, $f_{t-1}$ and $f_t$ are driven toward idempotence, feeding back $f_t$’s own output does not alter the prediction of $f_{t-1}$, yielding more reliable predictions across tasks and improving calibration in continual learning.
\end{itemize}
\section{$t$-SNE visualization of various methods}
In this section, we show more $t$-SNE visualization of various methods on first task testing data on CIFAR-100 with 500 buffer size. The task number is 10. $t$-SNE figures shows our method has the better capability of resisting catstrophic forgetting compared with ER, DER and BFP. We observed that the feature clusters of the 10 classes from the first task become increasingly blurred as the model learns knowledge from new tasks in these methods. However, this phenomenon is alleviated in our method.
\begin{figure}[H]
    \centering
    \includegraphics[width=0.8\linewidth]{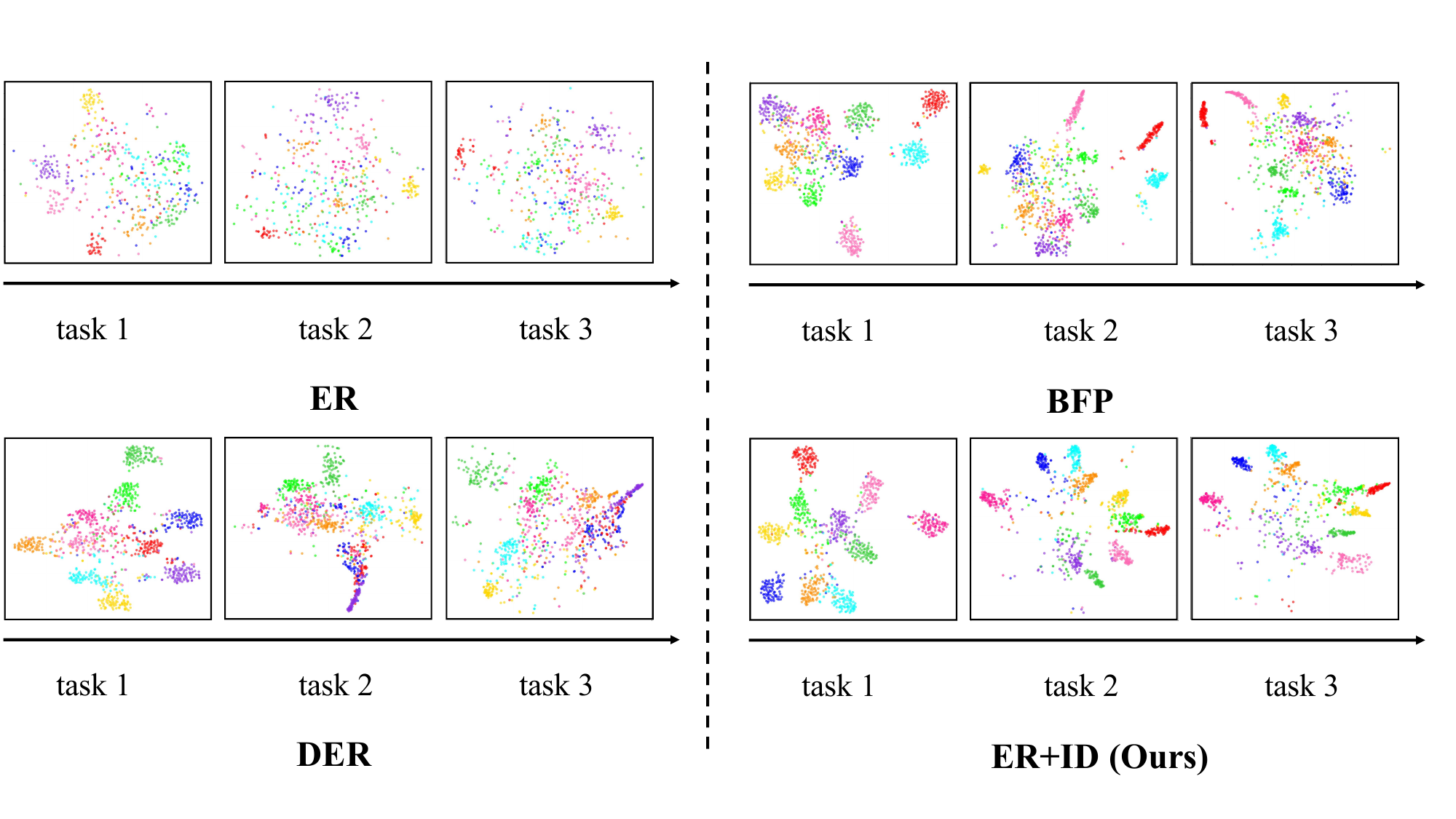}
    %\captionsetup{font=small}
    \caption{We perform $t$-SNE visualization of the features extracted from the first task testing data on CIFAR-100 across training tasks. The figures show how the feature clusters of the 10 classes from the first task change when the model train data from new tasks.}
    \label{fig:tsne}
\end{figure}
\begin{table}[t]
\centering
\captionsetup{font=small}
\caption{Comparison of Final Forgetting (FF) across different continual learning methods. All experiments are repeated 5 times with different seeds. The best results (lowest forgetting) are highlighted in \textcolor{myblue}{blue}. The second best results are highlighted in \textcolor{mygreen}{green}.}
\renewcommand{\arraystretch}{1.4}
\resizebox{0.9\textwidth}{!}{
\begin{tabular}{l cc cc cc}
\toprule
\multirow{2}{*}{\textbf{Method}} & \multicolumn{2}{c}{\textbf{CIFAR-10}} & \multicolumn{2}{c}{\textbf{CIFAR-100}} & \multicolumn{2}{c}{\textbf{Tiny-ImageNet}} \\
& Buffer 200 & Buffer 500 & Buffer 500 & Buffer 2000 & Buffer 500 & Buffer 4000 \\
\midrule
Joint & 0.00 & 0.00 & 0.00 & 0.00 & 0.00 & 0.00 \\
\midrule
iCaRL~\cite{rebuffi2017icarl} & 36.00\small{±7.29} & 30.19\small{±3.38} & 28.58\small{±0.84} & 24.24\small{±0.58} & 20.82\small{±1.31} & \textcolor{myblue}{15.29\small{±0.53}}\\
ER~\cite{riemer2018learning} & 71.35\small{±7.77} & 52.12\small{±7.56} & 71.92\small{±0.74} & 51.82\small{±0.75} & 74.79\small{±0.67} & 57.47\small{±0.57} \\
BiC~\cite{wu2019large} & 53.63\small{±7.18} & 24.87\small{±0.98} & 48.87\small{±0.91} & 38.50\small{±1.09} & 67.57\small{±1.98} & 63.48\small{±0.28} \\
LUCIR~\cite{hou2019learning} & 59.79\small{±10.16} & 37.58\small{±3.80} & 50.22\small{±1.26} & 32.48\small{±0.76} & 35.02\small{±0.56} & 30.59\small{±0.95} \\
DER~\cite{buzzega2020dark} & 45.31\small{±2.73} & 32.04\small{±2.88} & 56.66\small{±2.55} & 34.41\small{±2.05} & 68.43\small{±2.73}  & 59.54\small{±6.13} \\
DER++~\cite{buzzega2020dark} & 36.20\small{±4.15} & 27.93\small{±3.34} & 51.85\small{±1.61} & 34.44\small{±1.42} & 61.45\small{±3.12} & 39.11\small{±3.66} \\
ER-ACE~\cite{caccia2021new} & 19.52\small{±1.23} & 12.87\small{±1.06} & 38.61\small{±1.15} & 28.42\small{±0.55} & 40.97\small{±1.38} & 29.37\small{±1.09}\\
XDER~\cite{boschini2022class} & 16.36\small{±0.91} & 12.81\small{±0.48} & \textcolor{myblue}{24.15\small{±1.37}} & \textcolor{myblue}{11.17\small{±1.21}} & 42.90\small{±0.54} & 18.87\small{±1.08}\\
BFP~\cite{gu2023preserving} & 22.53\small{±5.00} & 16.81\small{±1.11} & 35.32\small{±3.94} & 19.76\small{±0.87} & \textcolor{mygreen}{30.02\small{±4.37}} & 27.19\small{±6.53}\\
\midrule
\textbf{Ours (ER+ID)} & \textcolor{myblue}{15.28\small{±2.41}} & \textcolor{myblue}{11.93\small{±0.49}} & 29.98\small{±2.52} & 17.46\small{±1.04} & 36.63\small{±3.37} & 22.46\small{±±1.86} \\
\textbf{Ours (BFP+ID)} & \textcolor{mygreen}{15.79\small{±2.73}} & \textcolor{mygreen}{12.11\small{±0.82}} & \textcolor{mygreen}{26.56\small{±2.56}} & \textcolor{mygreen}{12.52\small{±1.29}} & \textcolor{myblue}{27.41\small{±3.92}} & \textcolor{mygreen}{16.68\small{±0.44}} \\
\bottomrule
\end{tabular}%
}
\label{tab:forgetting_results}
\end{table}
\FloatBarrier
\vspace*{0.1\baselineskip}
\section{Forgetting comparison of different rehearsal-based continual learning methods}
Instead of FAA, Final Forgetting (FF) reflects the model's anti-forgetting capacity. To make fair comparison, we exclude the Exponential Moving Average (EMA) based methods, as the highest model performance on each task is from working model instead of EMA model, thereby reducing FF following Eq.~\ref{eq:FF} in appendix. FF measures the drop from each task’s historical peak accuracy (its best accuracy when first learned) to its final accuracy after all tasks, while EMA artificially smooths performance drops by reducing the accuracy it is first learned, thus understating true forgetting.% 只与增加的方法比
The Table~\ref{tab:forgetting_results} shows Final Forgetting (FF) of different continual learning methods. The results show that the idempotence loss yields a lower FF, indicating improved stability. It is worth noting that compared with XDER, there are no additional architectures or learnable parameters introduced in our method, just forwarding pass the model twice.

\section{Experimental Setting}
We evaluate our method on three standard continual learning benchmarks under Class-IL setting, where task identifiers are unavailable during testing, making it a challenging scenario for maintaining performance across tasks.

\noindent\textbf{Datasets.} Our experiments use three datasets with varying complexity:
\begin{itemize}
    \item \textbf{Split CIFAR-10}: The CIFAR-10 dataset is divided into 5 sequential tasks, each containing 2 classes. Each class comprises 5,000 training and 1,000 test images of size 32×32.
    \item \textbf{Split CIFAR-100}: CIFAR-100 is split into 10 tasks with 10 classes per task. Each class contains 500 training and 100 test images of size 32×32.
    \item \textbf{Split TinyImageNet}: TinyImageNet is divided into 10 tasks with 20 classes each. Each class has 500 training images, 50 validation images, and 50 test images.
\end{itemize}

\noindent\textbf{Evaluation Metrics.} We use two standard metrics to evaluate continual learning performance:
\begin{itemize}
    \item \textbf{Final Average Accuracy (FAA)}: Measures the average accuracy across all tasks after training is complete. For a model that has finished training on task $t$, let $a_i^t$ denote the test accuracy on task $i$. FAA is computed as the mean accuracy across all tasks.
    \item \textbf{Final Forgetting (FF)}: Quantifies how much knowledge of previous tasks is forgotten, defined as:
    \begin{equation}
        \text{FF} = \frac{1}{T-1}\sum_{i=1}^{T-1} \max_{j \in \{1,\cdots,T-1\}} (a_i^j - a_i^T).
        \label{eq:FF}
    \end{equation}
    where lower values indicate better retention of previously learned tasks.
    \item \textbf{ Expected Calibration Error (ECE)}: Quantifies the mismatch between a model's predicted confidence and its actual accuracy. Predictions are partitioned into $M$ confidence interval bins $B_m$. The ECE is computed as the weighted average of the absolute difference between the average confidence ($\mathrm{conf}(B_m)$) and the average accuracy ($\mathrm{acc}(B_m)$) within each bin:
    \begin{equation}
    \text{ECE} = \sum_{m=1}^{M} \frac{|B_m|}{N} \left| \mathrm{conf}(B_m) - \mathrm{acc}(B_m) \right|.
    \end{equation}
    where $N$ is the total number of samples. A lower ECE indicates a better-calibrated model whose confidence estimates are more reliable.

\end{itemize}
\subsection{Implementation Details supplementary}
%We follow ~\citet{gu2023preserving} using ResNet-18~\cite{he2016deep} as our backbone network for all experiments. This is the same
%base network used in rehearsal-based methods. All experiments
%are trained from scratch using SGD optimizer.
%with momentum 0 and weight decay 0. 
%The selected hyperparameters for each method are provided in the appendix following ~\citet{buzzega2020dark,gu2023preserving} . Note that we use uniform settings (epochs, batch size, memory batch size, and optimizer) across different methods to ensure fair comparison. In practice, we follow ~\cite{buzzega2020dark} by splitting CIFAR-10~\cite{krizhevsky2009learning} into 5 tasks and splitting CIFAR-100~\cite{krizhevsky2009learning} and TinyImageNet~\cite{wu2017tiny} into 10 tasks. 
Besides the details mentioned above, we train 50 epochs per task for Split CIFAR-10 and Split CIFAR-100 and 100 epochs per task for Split TinyImageNet~\cite{le2015tiny}. For Split CIFAR100, the learning rate is decreased by a factor of 0.1 at epochs 35 and 45, while for Split TinyImageNet,the learning rate is decreased by a factor of 0.1 at epochs 35, 60 and 75. The learning rate may vary in the light of different continual learning methods, while for a fair comparison, we use the same initial learning rate as DER and BFP for our methods. If not specified, all baselines use the reservoir sampling algorithm ~\cite{vitter1985random} to update memory, while BFP~\cite{gu2023preserving} uses class-balanced reservoir sampling ~\cite{buzzega2021rethinking} for pushing balanced examples into the buffer. 
\subsection{Hyperparameters}
In this section, we show hyperparameter combination that used in our experiments. These hyperparameters are
adopted from ~\citet{buzzega2020dark,boschini2022class,gu2023preserving} to make fair comparison.
\subsection*{Split CIFAR-10}

\textbf{Buffer size = 200}
\vspace{1em}
\begin{framed}
 \footnotesize
 \setlength{\parskip}{2pt}
 \setlength{\parindent}{0pt}
 \renewcommand{\arraystretch}{0.95}
 \textbf{iCaRL:} $lr = 0.1, \; wd = 10^{-5}$ 
 
 \textbf{LUCIR:} $\lambda_{\text{base}} = 5, \; \mathrm{mom} = 0.9, \; k = 2, \; \mathrm{epoch}_{\text{fitting}} = 20, \; lr = 0.03, \; lr_{\text{fitting}} = 0.01, \; m = 0.5$ 
 
 \textbf{BiC:} $\tau = 2, \; \mathrm{epochs}_{\mathrm{BiC}} = 250, \; lr = 0.03$
 
 \textbf{ER-ACE:} $lr = 0.03$ 
 
 \textbf{ER:} $lr = 0.1$ 
 
 \textbf{DER:} $lr = 0.03, \; \alpha = 0.3$  
 
 \textbf{DER++:} $lr = 0.03, \; \alpha = 0.1, \; \beta = 0.5$ 
 
 \textbf{XDER:}  $\alpha = 0.3,$ $m = 0.7, \; \beta = 0.9, \; \gamma = 0.85, \; wd = 1e-06, \; \lambda = 0.05, \; \eta = 0.001, \;  lr = 0.03, \;  \tau = 5, \; mom = 0.9$ 

 \textbf{DEP++ w/ BFP:} $lr = 0.03, \; \alpha_{distill} = 0.1, \; \alpha_{ce} = 0.5, \; \alpha_{bfp} = 1$
\end{framed}
\textbf{Buffer size = 500}
\vspace{1em}
\begin{framed}
 \footnotesize
 \setlength{\parskip}{2pt}
 \setlength{\parindent}{0pt}
 \renewcommand{\arraystretch}{0.95}
 \textbf{iCaRL:} $lr = 0.1, \; wd = 10^{-5}$
 
 \textbf{LUCIR:} $\lambda_{\text{base}} = 5, \; \mathrm{mom} = 0.9, \; k = 2, \; \mathrm{epoch}_{\text{fitting}} = 20, \; lr = 0.03, \; lr_{\text{fitting}} = 0.01, \; m = 0.5$
 
 \textbf{BiC:} $\tau = 2, \; 
 \mathrm{epochs}_{\mathrm{BiC}} = 250, \; lr = 0.03$
 
 \textbf{ER-ACE:} $lr = 0.03$
 
 \textbf{ER:} $lr = 0.1$
 
 \textbf{DER:} $lr = 0.03, \; \alpha = 0.3$
 
 \textbf{DER++:} $lr = 0.03, \; \alpha = 0.1, \; \beta = 0.5$
 
  \textbf{XDER:} $\alpha = 0.3,$ $m = 0.7,  \;  \; \beta = 0.9, \; \gamma = 0.85, \; wd = 1e-06, \; \lambda = 0.0, \; \eta = 0.001, \; lr = 0.03, \;  \tau = 5, \; mom = 0.9$
 
 \textbf{DEP++ w/ BFP:} $lr = 0.03, \; \alpha_{distill} = 0.2, \; \alpha_{ce} = 0.5, \; \alpha_{bfp} = 1$
\end{framed}
\subsection*{Split CIFAR-100}

\textbf{Buffer size = 500}
\vspace{1em}
\begin{framed}
 \footnotesize
 \setlength{\parskip}{2pt}
 \setlength{\parindent}{0pt}
 \renewcommand{\arraystretch}{0.95}
 \textbf{iCaRL:} $lr = 0.3, \; wd = 10^{-5}$
 
 \textbf{LUCIR:} $\lambda_{\text{base}} = 5, \; \mathrm{mom} = 0.9, \; k = 2, \; \mathrm{epoch}_{\text{fitting}} = 20, \; lr = 0.03, \; lr_{\text{fitting}} = 0.01, \; m = 0.5$
 
 \textbf{BiC:} $\tau = 2, \; \mathrm{epochs}_{\mathrm{BiC}} = 250, \; lr = 0.03$
 
 \textbf{ER-ACE:} $lr = 0.03$
 
 \textbf{ER:} $lr = 0.1$
 
 \textbf{DER:} $lr = 0.03, \; \alpha = 0.3$
 
 \textbf{DER++:} $lr = 0.03, \; \alpha = 0.1, \; \beta = 0.5$
 
 \textbf{XDER:} $\alpha = 0.3,$ $m = 0.7,  \; \beta = 0.9, \; \gamma = 0.85, \; wd = 1e-06, \; \lambda = 0.05, \; \eta = 0.001, \;  lr = 0.03, \;  \tau = 5, \; mom = 0.9$
 
 \textbf{DEP++ w/ BFP:} $lr = 0.03, \; \alpha_{distill} = 0.1, \; \alpha_{ce} = 0.5, \; \alpha_{bfp} = 1$
\end{framed}
\textbf{Buffer size = 2000}
\vspace{1em}
\begin{framed}
 \footnotesize
 \setlength{\parskip}{2pt}
 \setlength{\parindent}{0pt}
 \renewcommand{\arraystretch}{0.95}
 \textbf{iCaRL:} $lr = 0.3, \; wd = 10^{-5}$
 
 \textbf{LUCIR:} $\lambda_{\text{base}} = 5, \; \mathrm{mom} = 0.9, \; k = 2, \; \mathrm{epoch}_{\text{fitting}} = 20, \; lr = 0.03, \; lr_{\text{fitting}} = 0.01, \; m = 0.5$
 
\textbf{BiC:} $\tau = 2, \;
 \mathrm{epochs}_{\mathrm{BiC}} = 250, \; lr = 0.03$

\textbf{ER-ACE:} $lr = 0.03$

\textbf{ER:} $lr = 0.1$

\textbf{DER:} $lr = 0.03, \; \alpha = 0.3$

\textbf{DER++:} $lr = 0.03, \; \alpha = 0.1, \; \beta = 0.5$

\textbf{XDER:}  $\alpha = 0.3,$ $m = 0.7,  \; \beta = 0.9, \; \gamma = 0.85, \; wd = 1e-06, \; \lambda = 0.05, \; \eta = 0.001, \;  lr = 0.03, \;  \tau = 5, \; mom = 0.9$

\textbf{DEP++ w/ BFP:} $lr = 0.03, \; \alpha_{distill} = 0.1, \; \alpha_{ce} = 0.5, \; \alpha_{bfp} = 1$
\end{framed}
\subsection*{Split TinyImageNet}

\textbf{Buffer size = 4000}
\vspace{1em}
\begin{framed}
 \footnotesize
 \setlength{\parskip}{2pt}
 \setlength{\parindent}{0pt}
 \renewcommand{\arraystretch}{0.95}
 \textbf{iCaRL:} $lr = 0.03, \; wd = 10^{-5}$

\textbf{LUCIR:} $\lambda_{\text{base}} = 5, \; \mathrm{mom} = 0.9, \; k = 2, \; \mathrm{epoch}_{\text{fitting}} = 20, \; lr = 0.03, \; lr_{\text{fitting}} = 0.01, \; m = 0.5$

\textbf{BiC:} $\tau = 2, \; \mathrm{epochs}_{\mathrm{BiC}} = 250, \; lr = 0.03$

\textbf{ER-ACE:} $lr = 0.03$

\textbf{ER:} $lr = 0.1$

\textbf{DER:} $lr = 0.03, \; \alpha = 0.1$

\textbf{DER++:} $lr = 0.03, \; \alpha = 0.1, \; \beta = 0.5$

\textbf{XDER:}  $\alpha = 0.3,$ $m = 0.7,  \; \beta = 0.9, \; \gamma = 0.85, \; wd = 1e-06, \; \lambda = 0.0, \; \eta = 0.001, \;  lr = 0.03, \;  \tau = 5, \; mom = 0.9$

\textbf{DEP++ w/ BFP:} $lr = 0.03, \; \alpha_{distill} = 0.3, \; \alpha_{ce} = 0.8, \; \alpha_{bfp} = 1$
\end{framed}

\subsection{Intergrated IDER into BFP}
As ~\citet{gu2023preserving} introduces BFP distillation loss, which focuses on features. We can easily incorporate our method into BFP framework.
The BFP loss is:
\begin{equation}
    \mathcal{L}_{BFP} = \sum_{(x, y) \in \mathcal{T}_{t},M} \| Ah_{t}(x,0) - h_{t-1}(x,0) \|_2,
\end{equation} 
where $h_{t}$ is feature extractor in model $f_t$ on the t-th task and $A$ is linear transformation aimed to preserve the linear
separability of features backward in time.

During training on the task $t$, the model $f_t$ and $A$ are optimized respectively.
Thus, the $\mathcal{L}_{BFP+ID}$ can be:
\begin{equation}
    \mathcal{L}_{BFP+ID} = \mathcal{L}_{ice} + \alpha \mathcal{L}_{ide} + \beta \mathcal{L}_{rep\text{-}ice}+ \gamma \mathcal{L}_{BFP}.
\end{equation} 
\subsection{Complexity and Training Cost}
We train all experiments on NVIDIA RTX 4090 and Ascend 910B. The additional parameters we need for modified architecture are very small. Using ResNet-18 as the backbone, the normal architecture contains 11.22M parameters on CIFAR-100 while the modified architecture increases the parameter count to 11.91M parameters. Although we need two forward passes to train the model, which leads to a slightly longer training time compared with DER++, the longer training time is acceptable as it increases the performance by a significant margin.

\section{Limitations}

\noindent\textbf{Naive Implementation.} As we first introduce idempotence property into continual learning, the method should be very simple. In the future, we try to combine our method with more complementary techniques to further improve performance. In addition, we also plan to explore the application of the idempotence property in data sampling strategies for continual learning.

\section{ Ablation Results}

\subsection{contribution of each component}
We conduct a component ablation study to isolate the contributions of each part of the overall objective: the Standard Idempotent Module (SIM), the Idempotent Distillation Module (IDM), and Experience Replay (ER). We focus on the ER+ID method for this study. The ablation study results are shown in Table~\ref{tab:ablation_c}. Using only the Standard Idempotent Module (SIM) or combing SIM with ER produces similar performance compared with finetuning or ER baseline,
\begin{wraptable}{r}{0.4\textwidth}
\centering
\vspace{-10pt}  
%\captionsetup{font={small}}
\caption{Ablation study of different components on Split CIFAR-100.}

\label{tab:ablation_c}
\renewcommand{\arraystretch}{1.3}
\begin{tabular}{cccc}
\toprule
\textbf{SIM} & \textbf{IDM} & \textbf{ER} & \textbf{FAA} \\
\midrule
\ding{51} & \ding{55} & \ding{55} & 8.23 \\
\ding{51} & \ding{55} & \ding{51} & 24.73 \\
\ding{51} & \ding{51} & \ding{51} & 44.82 \\
\bottomrule
\end{tabular}

\end{wraptable}
 indicating that SIM alone trains model to be idempotent on the current task, which making it well-suited for subsequent idempotent distillation and  doesn't mitigate catastrophic forgetting. It also demonstrates that modified architecture does not influence performance and the observed improvements benefit from idempotent distillation loss.  Whta's more, adding the Idempotent Distillation Module (IDM) yields substantial performance gains, which further improves accuracy to 44.82$\%$. This validates the effectiveness of idempotence distillation loss.
\subsection{Hyperparameter sensitivity}
We performed ablations on CIFAR-100 with 500 buffer size under CIL setting. We ablate on ER+ID method, as it consistently yields substantial improvements over ER baseline across all datasets and does not introduce any additional hyperparameters. Both $\alpha$ and $\beta$ are from the set $\{ 0.1, 0.2, 0.5, 1\}$. Table~\ref{tab:hyperparameters} demonstrates that IDER is not overly sensitive to specific hyperparameter values, as multiple configurations yield consistent performance improvements. This reliability underscores IDER’s practicality and suitability for a wide range of continual learning applications. 
\begin{comment}
\begin{table}[H]
\centering
\captionsetup{font={small}}
\caption{Impact of hyperparameters $\alpha$ and $\beta$ for ER+ID on Final Average Accuracy (FAA) on Split CIFAR-100. We use learning rate as 0.03,training epoch as 50.}
\label{tab:hyperparameters}
\renewcommand{\arraystretch}{1.3}
\begin{tabular}{ccc}
\toprule
\textbf{alpha} & \textbf{beta} & \textbf{FAA} \\
\midrule
\multirow{4}{*}{0.1} & 0.1 & 44.77 \\
                     & 0.2 & 44.41 \\
                     & 0.5 & 43.82 \\
                     & 1   & 41.20 \\
\cmidrule(lr){1-3}
\multirow{4}{*}{0.2} & 0.1 & 44.24 \\
                     & 0.2 & 44.43 \\
                     & 0.5 & 44.33 \\
                     & 1   & 41.58 \\
\cmidrule(lr){1-3}
\multirow{4}{*}{0.5} & 0.1 & 41.74 \\
                     & 0.2 & 44.25 \\
                     & 0.5 & \textbf{44.82} \\
                     & 1   & 42.72 \\
\cmidrule(lr){1-3}
\multirow{4}{*}{1}   & 0.1 & 40.42 \\
                     & 0.2 & 43.02 \\
                     & 0.5 & 43.87 \\
                     & 1   & 40.93 \\
\bottomrule
\end{tabular}
\label{tab:comparison_p}
\end{table}
\end{comment}
\begin{table}[H]
\centering
%\captionsetup{font={small}}
\caption{Impact of hyperparameters $\alpha$ and $\beta$ for ER+ID on Final Average Accuracy (FAA) on Split CIFAR-100. We use learning rate as 0.03 and training epoch as 50.}
\label{tab:hyperparameters}
\renewcommand{\arraystretch}{1.3}
\begin{tabular}{c cccc}
\toprule
\textbf{beta} & \textbf{alpha=0.1} & \textbf{alpha=0.2} & \textbf{alpha=0.5} & \textbf{alpha=1} \\
\midrule
0.1 & 44.77 & 44.24 & 41.74 & 40.42 \\
0.2 & 44.41 & 44.43 & 44.25 & 43.02 \\
0.5 & 43.82 & 44.33 & \textbf{44.82} & 43.87 \\
1   & 41.20 & 41.58 & 42.72 & 40.93 \\
\bottomrule
\end{tabular}
\end{table}

\subsection{Probability selection}
We provide ablation study of the probability $P$ used in the Standard Idempotent Module, which determines whether the second input is set to the empty signal  or the ground-truth one-hot vector. We perform a sensitivity ablation on Split CIFAR-100 with 500 buffer size and on ER+ID method. In our ablation study, the second input is set to the neutral ``empty'' signal input 0 with probability $P$ and the ground-truth one-hot vector y with probability $1-P$. Table~\ref{tab:hyperparameter_p} shows that as $P$ increases, FAA consistently improves and peaks at $P=0.9$, which reaches 44.82. Then there is a slight drop at $P=1.0$, where FAA is 43.26. Therefore, we use $P=0.9$ by default. For simplicity, we can also set $P=1.0$, which yields comparable performance.%Given that, we choose $P=0.9$ as a default in our method. You can also choose $P=1.0$ as more simple method which achieves comparable performance.
\begin{table}[H]
\centering
%\captionsetup{font={small,color=blue},labelfont={color=blue}}
%\captionsetup{font={small}}
\caption{Impact of the probability $P$ on Final Average Accuracy (FAA) on Split CIFAR-100.}

\label{tab:hyperparameter_p}
\renewcommand{\arraystretch}{1.3}
\begin{tabular}{cccccccccc}
\toprule
\textbf{P} & 0.2 & 0.3 & 0.4 & 0.5 & 0.6 & 0.7 & 0.8 & 0.9 & 1 \\
\midrule
\textbf{FAA} & 20.9 & 22.24 & 27.03 & 29.19 & 31.46 & 37.48 & 42.24 & \textbf{44.82} & 43.26 \\
\bottomrule
\end{tabular}

\end{table}

\section{Statement on the Use of Large Language Models}
We hereby declare that large language models (LLMs), specifically GPT-5 , were used during the preparation of this manuscript. The use of LLMs was strictly limited to:
aiding and polishing writing. The LLM was used solely as an assistive tool for prose refinement and did not contribute to the intellectual content of the research.

\end{document}